\documentclass{article} 
\usepackage{iclr2027_conference,times}

\usepackage{amsmath,amsfonts,bm}

\def\eqref#1{equation~\ref{#1}}

\def\1{\bm{1}}

\DeclareMathAlphabet{\mathsfit}{\encodingdefault}{\sfdefault}{m}{sl}
\SetMathAlphabet{\mathsfit}{bold}{\encodingdefault}{\sfdefault}{bx}{n}

\usepackage{amsmath,amssymb,amsthm}
\usepackage{booktabs}
\usepackage{xurl}
\usepackage{hyperref}
\usepackage{url}
\usepackage{graphicx}
\usepackage{xcolor}
\usepackage{colortbl}
\usepackage{array}
\usepackage{colortbl}
\definecolor{mygray}{HTML}{F2F2F2}
\definecolor{myblue}{HTML}{EAF2F8}
\usepackage{microtype}
\usepackage[most]{tcolorbox}
\usepackage{subcaption}
\usepackage{wrapfig}
\usepackage{xspace}

\definecolor{Burgundy}{HTML}{9E2E2F}
\definecolor{SoftGray}{HTML}{F5F5F5}
\definecolor{BaseGray}{HTML}{EFEFED}
\usepackage{algorithm}
\usepackage{algpseudocode}
\definecolor{TakeawayBorder}{HTML}{9A4A50}
\definecolor{TakeawayBg}{HTML}{FCF8F7}
\definecolor{TakeawayTitle}{HTML}{7A2630}
\tcbset{
  takeaway/.style={
    enhanced,
    colback=TakeawayBg,
    colframe=TakeawayBorder,
    arc=2.5mm,
    boxrule=0.65pt,
    left=3.5mm,
    right=3.5mm,
    top=3.2mm,
    bottom=2.6mm,
    before skip=3mm,
    after skip=3mm,
    fonttitle=\bfseries,
    coltitle=white,
    attach boxed title to top left={
      xshift=3.5mm,
      yshift=-2.2mm
    },
    boxed title style={
      colback=TakeawayTitle,
      colframe=TakeawayTitle,
      boxrule=0pt,
      arc=1.7mm,
      left=3mm,
      right=3mm,
      top=0.8mm,
      bottom=0.8mm
    }
  }
}
\iclrfinalcopy 

\definecolor{LinkRed}{HTML}{9E2E2F}      
\definecolor{TitleRed}{HTML}{7A2630}     
\definecolor{SectionRed}{HTML}{9E2E2F}   
\definecolor{CardBg}{HTML}{FCF8F7}       
\definecolor{CardFrame}{HTML}{DCC8C9}    
\definecolor{HeadGray}{HTML}{3D3D3D}     
\definecolor{HeadRule}{HTML}{4A4A4A}     
\definecolor{MetaGray}{HTML}{6B6B6B}     

\usepackage{longtable}
\newcommand{\git}{\raisebox{-1.5pt}{\includegraphics[height=1.05em]{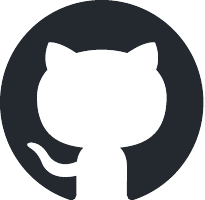}}\xspace}

\hypersetup{
    colorlinks=true,
    citecolor=LinkRed,
    linkcolor=LinkRed,
    urlcolor=LinkRed,
    pdftitle={What Does Multi-Turn Reinforcement Learning Really Improve in LLM Agents?},
    pdfauthor={Kejian Zhu, Zhuoran Jin, Shangqing Tu, Tianyi Men, Yupu Hao, Jiachun Li, Yuanming Zhang, Kang Liu, Jun Zhao}
}

\makeatletter
\def\section{\@startsection{section}{1}{\z@}{-2.0ex plus
    -0.5ex minus -.2ex}{1.5ex plus 0.3ex
minus0.2ex}{\large\sc\raggedright\color{SectionRed}}}
\def\subsection{\@startsection{subsection}{2}{\z@}{-1.8ex plus
-0.5ex minus -.2ex}{0.8ex plus .2ex}{\normalsize\sc\raggedright\color{SectionRed}}}
\def\subsubsection{\@startsection{subsubsection}{3}{\z@}{-1.5ex
plus      -0.5ex minus -.2ex}{0.5ex plus
.2ex}{\normalsize\sc\raggedright\color{SectionRed}}}
\makeatother

\newcommand{\RunningHead}{Rethinking Agentic RL}
\newcommand{\PaperDate}{November 2026}

\fancypagestyle{arxivfirst}{
  \fancyhead{}
  \fancyhead[L]{\small\bfseries\color{Burgundy}\RunningHead}
  \fancyhead[R]{\small\color{MetaGray}\PaperDate}
  \fancyfoot[C]{\small\color{MetaGray}\thepage}
  \renewcommand{\headrulewidth}{1.1pt}
  \renewcommand{\headrule}{{\color{HeadRule}\hrule width\headwidth height 1.1pt}}
}
\fancypagestyle{arxivrunning}{
  \fancyhead{}
  \fancyhead[C]{\small\color{HeadGray}\RunningHead}
  \fancyfoot[C]{\small\color{MetaGray}\thepage}
  \renewcommand{\headrulewidth}{1.1pt}
  \renewcommand{\headrule}{{\color{HeadRule}\hrule width\headwidth height 1.1pt}}
}
\tcbset{
  titlecard/.style={
    enhanced,
    colback=CardBg,
    colframe=CardFrame,
    boxrule=0.7pt,
    arc=2.8mm,
    left=4mm,
    right=4mm,
    top=4.5mm,
    bottom=3.5mm,
    before skip=1mm,
    after skip=6mm
  }
}
\title{What Does Multi-Turn Reinforcement Learning Really Improve in LLM Agents?}
\begin{document}

\thispagestyle{arxivfirst}

\begin{tcolorbox}[titlecard]
\begin{center}
  {\LARGE\bfseries\color{TitleRed}
   What Does Multi-Turn Reinforcement Learning\\[3pt]
   Really Improve in LLM Agents?\par}
  \vspace{10pt}
  {\bfseries Kejian Zhu\textsuperscript{1,2},
  Zhuoran Jin\textsuperscript{1,2},
  Shangqing Tu\textsuperscript{3},
  Tianyi Men\textsuperscript{1,2}\par}
  \vspace{2.5pt}
  {\bfseries Yupu Hao\textsuperscript{1,2},
  Jiachun Li\textsuperscript{1,2},
  Kang Liu\textsuperscript{1,2},
  Jun Zhao\textsuperscript{1,2}\par}
  \vspace{7pt}
  {\fontsize{9}{11.5}\selectfont
  \textsuperscript{1}The Key Laboratory of Cognition and Decision Intelligence for Complex Systems,\\
  Institute of Automation, Chinese Academy of Sciences, Beijing, China\\[1.5pt]
  \textsuperscript{2}School of Artificial Intelligence, University of Chinese Academy of Sciences
  \ \textsuperscript{3}Tsinghua University\par}
  \vspace{6pt}
  {\fontsize{9}{11.5}\selectfont\ttfamily\color{HeadGray}
  zhukejian2025@ia.ac.cn\quad
  \{zhuoran.jin, kliu, jzhao\}@nlpr.ia.ac.cn\par}
\end{center}
\vspace{3pt}
\begin{center}
  \textcolor{CardFrame}{\rule{0.94\linewidth}{0.5pt}}
\end{center}
\vspace{1pt}
\begin{center}
  {\large\sc\color{TitleRed} Abstract}
\end{center}
\vspace{-1pt}
While multi-turn agentic reinforcement learning (RL) excels on long-horizon tasks, its performance gains remain highly sensitive to interaction constraints. Evaluating across sampling breadth $K$ and interaction depth $T$ via $\mathrm{Pass}@(K,T)$, we show that trajectory-level RL advantages depend on sufficient turn budgets, diminishing significantly when budgets tighten. Besides, agentic RL does not uniformly elevate single-step decision quality; rather, its strength concentrates in error recovery and long-horizon resilience. To mitigate the deployment risks caused by this budget dependence, we provide a preliminary discussion on the benefits and limitations of process supervision signals and introduce BudgetMix RL as an alternative that varies interaction budgets during training. Experiments demonstrate that BudgetMix RL achieves robust performance across diverse execution settings.

\vspace{6pt}
{\small
\urlstyle{tt}
\begin{center}
    \renewcommand{\arraystretch}{1.3}
    \begin{tabular}{rcl}
          \git & \textbf{Project} & \url{https://rethinking-agentic-rl.github.io/}
    \end{tabular}
\end{center}
}
\end{tcolorbox}


\section{Introduction}

Multi-turn agentic reinforcement learning (RL) has recently emerged as an important paradigm for improving the long-horizon capabilities of large language model (LLM) agents \citep{Agent_Environment_Survey, agentgymrl}. By interacting with environments persistently, consuming feedback, and adapting subsequent actions, agents trained with RL have achieved substantial gains on multi-turn tasks (e.g., code or tool use). Despite these improvements, a fundamental question remains insufficiently understood: \textbf{what capabilities does multi-turn agentic RL actually improve?} Answering this question is important not only for understanding how agentic RL differs from its base model, but also for assessing whether agentic RL models remain reliable under realistic deployment scenarios.


Existing works primarily evaluate agents at the trajectory level, typically based on whether a task is completed within a given budget. Prior work on single-turn RL further uses \textbf{$\mathrm{Pass}@K$} to probe model capability boundary \citep{limitrlvr}, which increases the number of independently sampled solutions. $\mathrm{Pass}@K$ captures one dimension of test-time scaling: \emph{sampling breadth}. Thus, we first apply $\mathrm{Pass}@K$ to the evaluation of agents in \S\ref{sec:trajectory}. The results show that agentic RL models consistently outperform base models as the number of sampled trajectories $k$ increases. These results suggest that agentic RL effectively expands the empirical capability boundary of the base model.

\begin{figure}[t]
\centering
\includegraphics[width=\textwidth]{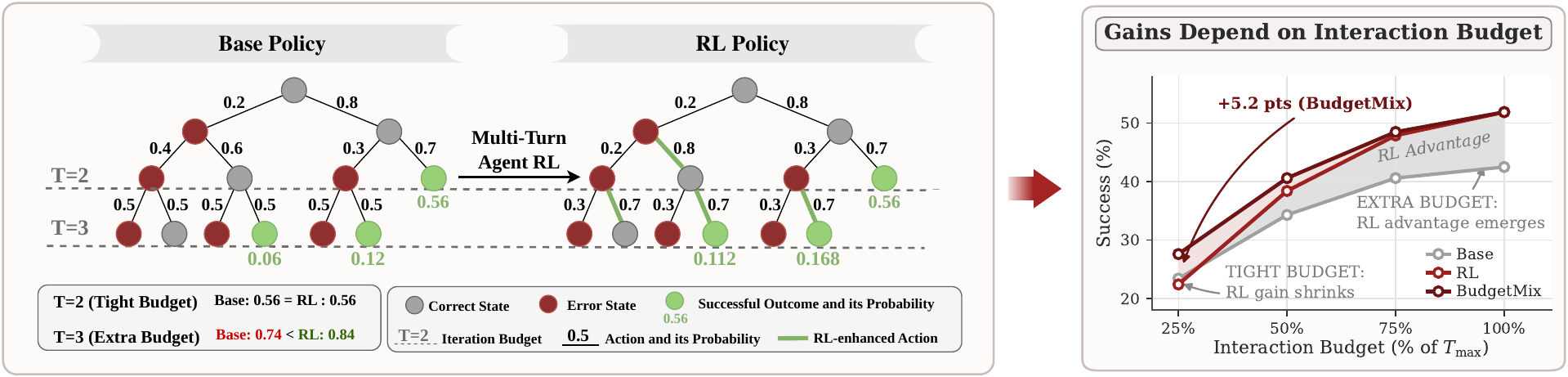}
\caption{\textbf{Overview of our findings.} \textbf{Left:} Multi-turn agentic RL does not uniformly enhance per-step action quality; rather, it mainly enhances ability of error recovery and long-horizon resilience. \textbf{Right:} The trajectory-level gains of agentic RL depend on a sufficient interaction budget ($T$).}
\label{fig:intro}
\vspace{-5pt}
\end{figure}

However, for interactive agents, there is another test-time dimension: the number of interactions allowed per trajectory \citep{shen2026thinking}, which captures \emph{interaction depth}. Accordingly, we evaluate performance using $\mathrm{Pass}@T$ by varying the maximum allowed interaction turns (\S\ref{sec:budget}). When we vary budget $T$, an interesting phenomenon emerges: given a large budget, agentic RL models demonstrate a substantial advantage over base models. However, under tighter budgets, this advantage shrinks considerably and can even disappear or reverse on certain tasks (right-hand panel of Figure~\ref{fig:intro}). These results suggest that the gains of \textbf{agentic RL depend on a sufficient interaction budget}. To ensure that this phenomenon is not merely an artifact of restrictive budgets rendering some tasks structurally unsolvable, we compute structural performance upper bounds and discuss in Appendix~\ref{app:pass_T_metrics}.

To further study this budget-dependent shift in capability, we analyze the step-level action quality of models before and after multi-turn RL (\S\ref{sec:step}). Results show that, given the same state and history, RL models do not consistently produce better next-step actions and can even perform worse in certain settings. Instead, their advantage becomes more pronounced at error states and later stages of a trajectory. As illustrated in Figure~\ref{fig:intro}, multi-turn RL does not uniformly improve action quality of every step; it primarily strengthens ability of error recovery and sustain long-horizon execution, which may partly explain why RL models need a larger budget $T$ to demonstrate their advantages.

To understand this phenomenon more deeply, we perform a series of mechanistic analyses (\S\ref{sec:mechanism}). Our log-likelihood analysis (\S\ref{sec:likelihood}) shows that RL-generated actions remain locally plausible under the base policy, yet complete RL trajectories receive substantially higher likelihood under the RL policy than under the base policy. This suggests that multi-turn RL may improve how reliably the model composes locally plausible actions into long-horizon trajectories. This phenomenon offers a possible explanation for how RL models can surpass the empirical $\mathrm{Pass}@K$ capability boundary of base models, even without consistently superior single-step actions to the base models. The training dynamics in \S\ref{sec:dynamics} further show how this pattern emerges progressively during training.


These analyses reveal a hidden risk masked by conventional trajectory-level success rates: models trained with multi-turn RL may fail to perform consistently under varying interaction budgets. Furthermore, their single-step action quality can decline, which are critical in real-world deployments involving strict budget limits or irreversible errors \citep{ma2026timely,doshi2026towards}. To address this, we provide a preliminary discussion on some candidate solutions (\S\ref{sec:method}). Specifically, we examine the benefits and limitations of process supervision signals and introduce \textbf{BudgetMix RL}, an alternative training strategy that varies interaction budgets. Experiments show that BudgetMix RL consistently improves performance across budgets while mitigating the decline in step-level action quality (\S\ref{sec:budgetmix}).

\section{Preliminaries and Evaluation Framework}

\label{sec:preliminaries}
This section formalizes our research setting. We first describe multi-turn agentic reinforcement learning (\S\ref{sec:setting_multi_turn_rl}) and then present the two-dimensional $\mathrm{Pass}@(K,T)$ evaluation metric (\S\ref{sec:pass_kt}).

\subsection{Multi-Turn Agentic Reinforcement Learning}
\label{sec:setting_multi_turn_rl}
\vspace{-5pt}
We consider an agent interacting with an environment over multiple turns to solve a task $x \sim \mathcal{D}$. At step $t$, given the interaction history $h_t = (x, o_1, a_1, \ldots, o_{t-1}, a_{t-1}, o_t)$, the agent samples an action $a_t \sim \pi(\cdot \mid h_t)$, prompting the environment to return observation $o_{t+1} \sim P(\cdot \mid h_t, a_t)$:
\begin{equation*}
a_t \sim \pi(\cdot \mid h_t),
\qquad
o_{t+1} \sim P(\cdot \mid h_t, a_t).
\end{equation*}
Actions can represent tool calls, terminal operations or code execution. Under an interaction budget of at most $T$ turns, an episode yields a trajectory $\tau_{\leq T} = (x, o_1, a_1, \ldots, o_T, a_T)$, scored by a terminal verifier with reward $R(\tau_{\leq T})$. Our framework accommodates both outcome-only RL (e.g., GRPO \citep{guo2025deepseek}) and process-level credit assignment methods (e.g., GiGPO \citep{gigpo}).

\subsection{\texorpdfstring{$\mathrm{Pass}@(K,T)$}{Pass@(K,T)}: Sampling Breadth and Interaction Depth}
\label{sec:pass_kt}
Conventional agent evaluations typically report trajectory success. To probe capability boundaries, prior work on single-turn RL uses the metric $\mathrm{Pass}@K$, the probability that at least one of $k$ independent samples succeeds, which can be seen as \emph{sampling breadth}. Interactive agents introduce an orthogonal axis: the maximum interaction budget $T$ per trajectory, which governs \emph{interaction depth}. To capture both dimensions, we use metric $\mathrm{Pass}@(K,T)$ \citep{doesrlexpandcapability} for a policy $\pi$ on task $x$:
\begin{equation}
\label{eq:passkt}
\mathrm{Pass}@(K,T) = \Pr_{\tau^{(1)}, \dots, \tau^{(K)} \sim \pi(\cdot \mid x)} \Bigl[ \exists i \in \{1, \ldots, K\} : R(\tau^{(i)}_{\leq T}) = 1 \Bigr],
\end{equation}
which underpins our analysis in \S\ref{sec:analysis}. Fixing the sampling budget to a constant $C$ while sweeping $T$ yields $\mathrm{Pass}@T = \mathrm{Pass}@(K=C, T)$, isolating the depth axis evaluated in \S\ref{sec:budget}.

\section{When Does agentic RL Outperform the Base Model?}
\label{sec:analysis}

We now examine whether the gains from multi-turn agentic RL can be consistently achieved, as well as its potential risks. We evaluate models with two dimensional metrics $\mathrm{Pass}@(K,T)$ introduced in \S\ref{sec:pass_kt} (\S\ref{sec:trajectory} and \S\ref{sec:budget}), analyze step-level action quality (\S\ref{sec:step}), and summarize the potential risks (\S\ref{sec:risk}). We adopt GRPO as our primary RL algorithm and evaluate on four interactive benchmarks: DeepCoder-MT, BFCL, InterCode-SQL, and FrozenLake, spanning diverse domains and modalities. Additional results are reported in Appendix~\ref{app:additional_pass_k} and \ref{app:additional_pass_t}, implementation details in Appendix~\ref{app:training_details} and \ref{app:evaluation_details}.

\subsection{Trajectory-Level Gains across Sampling Breadth}
\label{sec:trajectory}

\begin{figure}[t]
\centering
\includegraphics[width=0.95\textwidth]{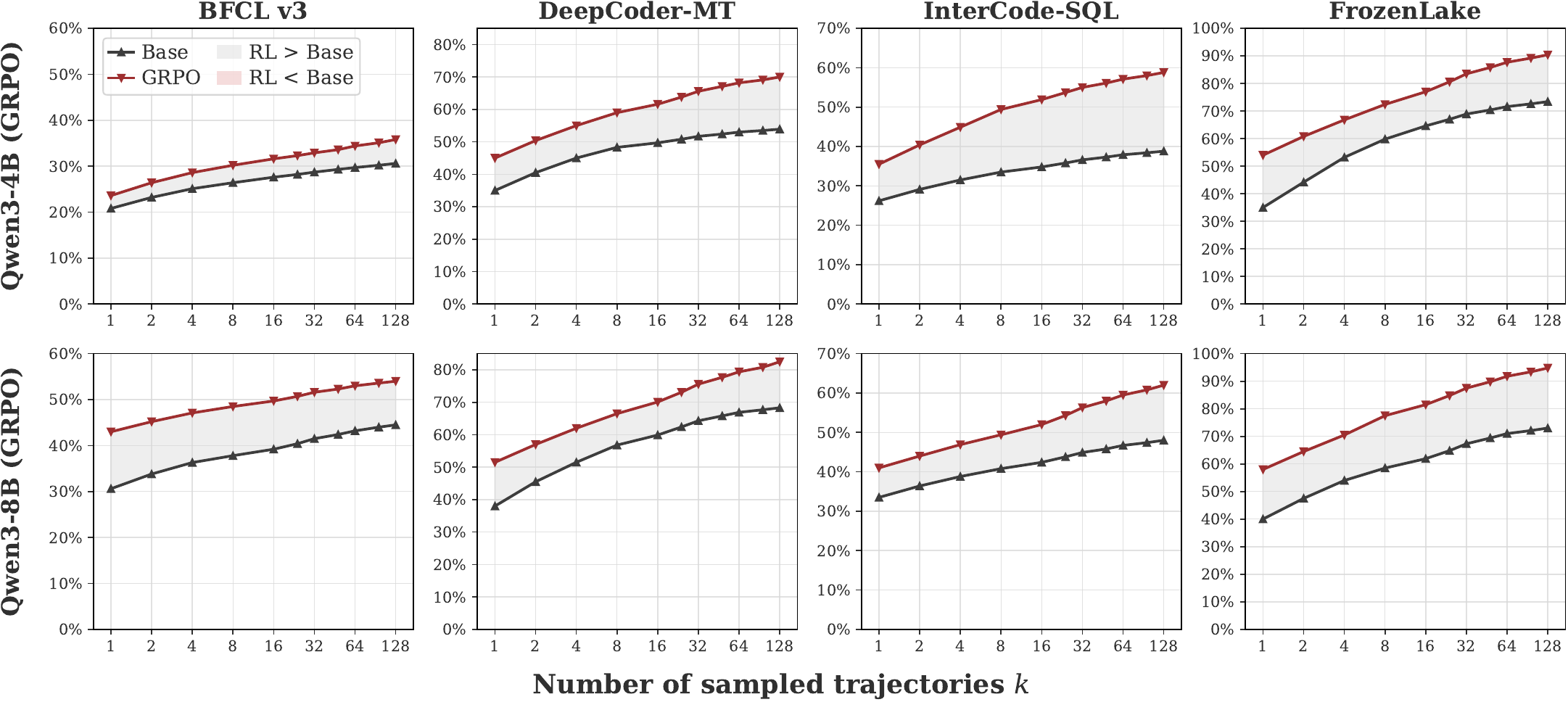}
\caption{$\mathrm{Pass}@K$ across sampling breadth. Sweeping the number of sampled trajectories $K$.}
\label{fig:passk}
\vspace{-10pt}
\end{figure}

We begin by evaluating the trajectory-level performance using the $\mathrm{Pass}@K$. By increasing the number of independently sampled trajectories, $\mathrm{Pass}@K$ probes the empirical capability boundary of a model, which we use to characterize how this boundary changes after multi-turn RL. As shown in Figure~\ref{fig:passk}, the RL model consistently outperforms the base model as the sampling breadth $k$ increases, suggesting that multi-turn RL can push trajectory-level success beyond the empirical capability boundary of base model. This contrasts with prior observations in single-turn tasks, where RL does not expand capability boundary beyond the base model. Further results are shown in Appendix~\ref{app:additional_pass_k}.

\vspace{-2pt}
\begin{tcolorbox}[takeaway,title={Takeaway 3.1}]
\textbf{Pros of Agentic RL:} unlike single-turn RL, multi-turn agentic RL can push trajectory-level performance beyond the empirical $\mathrm{Pass}@K$ capability boundary of the base model.
\end{tcolorbox}
\vspace{-5pt}

\subsection{RL Gains Shrink under Tight Interaction Budgets}
\label{sec:budget}

\begin{figure}[t]
\centering
\includegraphics[width=0.95\textwidth]{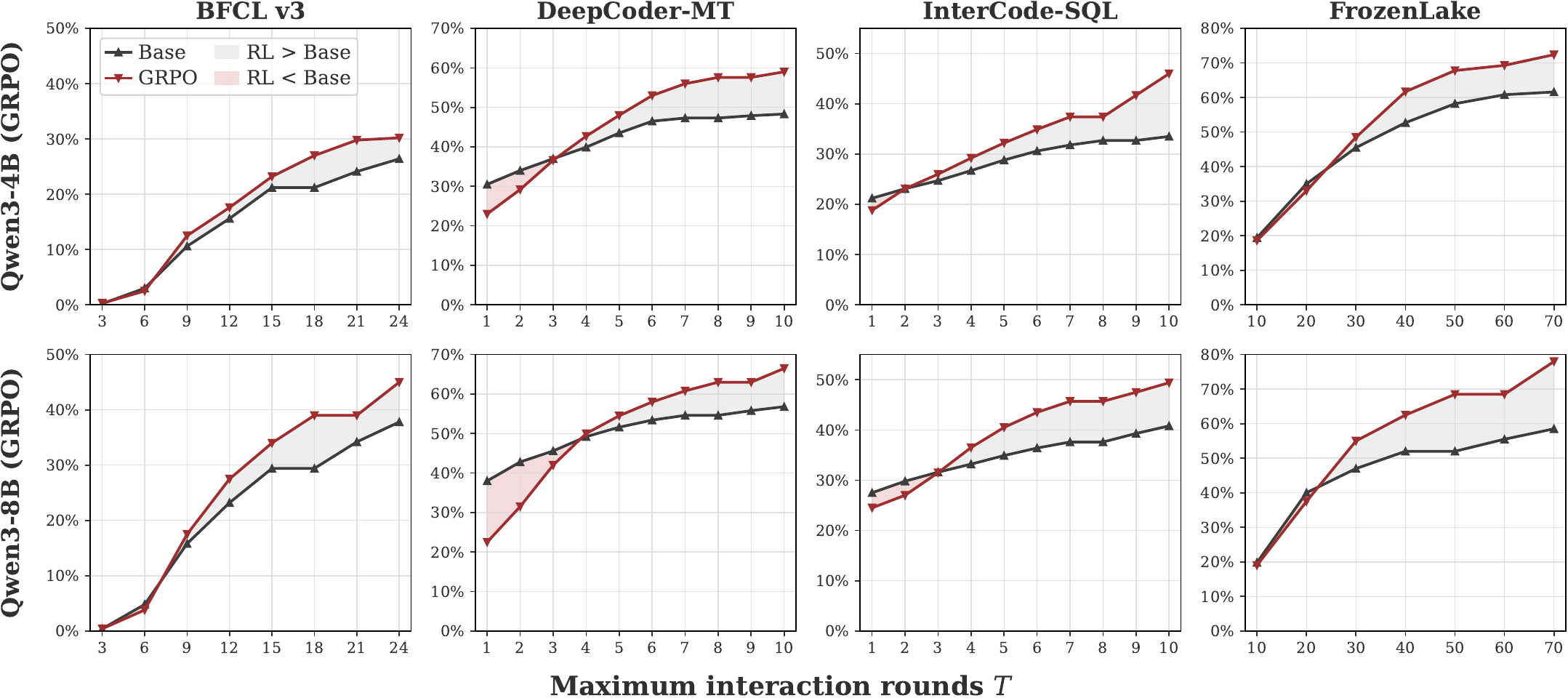}
\caption{$\mathrm{Pass}@T$ at fixed $k{=}8$. Sweeping the maximum number of interaction turns $T$.}
\label{fig:passt}
\vspace{-5pt}
\end{figure}

Beyond \textit{sampling breadth}, we evaluate how model performance evolves with \textit{interaction depth}. As shown in Figure~\ref{fig:passt}, we fix the sampling breadth at $k=8$ and vary the maximum interaction turns $T$. The results reveal an interesting pattern: \textbf{the advantage of the RL model depends on a sufficient interaction budget}. As the budget $T$ increases, the performance advantage of RL model over base model becomes increasingly pronounced. Conversely, under tight budgets (small $T$), this advantage diminishes, and can even reverse on certain tasks. To rule out the possibility that the performance drop at small $T$ is trivially caused by tasks becoming structurally unsolvable under restrictive budgets, we also compute the structural performance upper bound for each task and discuss in Appendix~\ref{app:pass_T_metrics}. Notably, at small $T$, even when substantial headroom remains below this upper bound, agentic RL fails to exhibit any clear advantage over the base model. More results are shown in Appendix~\ref{app:additional_pass_t}.

We conjecture that agent RL primarily strengthens long-horizon capabilities whose benefits emerge when sufficient interaction budget is available. Under restricted budgets, these benefits may disappear, suggesting that its gains \textbf{do not necessarily stem from uniformly better step-level actions}. 

\vspace{-2pt}
\begin{tcolorbox}[takeaway,title={Takeaway 3.2}]
\textbf{Potential Cons of Agentic RL:} the performance \textbf{gains from multi-turn RL depend on sufficient interaction budgets}, and can diminish or even reverse under tight budgets.
\end{tcolorbox}
\vspace{-5pt}

\subsection{Does Multi-Turn RL Improve Step-Level Action Quality?}
\label{sec:step}


The phenomenon of last section motivates a question: \textbf{does multi-turn RL improve action quality for each individual step?} Thus we compare the step quality of the base and RL model under identical states. For each state $s$, we independently sampled $G=16$ actions from both policies and averaged their quality scores to reduce sampling variance and obtain a more robust estimate of action quality:
\begin{equation*}
a_i^{\mathrm{Base}} \sim \pi_{\mathrm{Base}}(\cdot \mid s),
\qquad
a_i^{\mathrm{RL}} \sim \pi_{\mathrm{RL}}(\cdot \mid s),
\qquad i=1,\ldots,G.
\end{equation*} 
Strictly evaluating one-step quality is challenging because reliable process rewards remain difficult to obtain in agentic settings \citep{li2026verifiedcriticalstepoptimization}. We therefore focus on two tasks with reliable step-level evaluation: (1) For FrozenLake, we use finite-horizon dynamic programming (DP) with a discount factor $\gamma$ of \(0.95\) to compute the exact \(Q^*\) value of each step-level action under optimal continuation within the remaining interaction budget. (2) For DeepCoder-MT, the code of each step can be scored by unit-test pass rates. To isolate action quality from state visitation bias, we construct a state evaluation pool by sampling intermediate states equally from base and RL trajectories. Following experiments use Qwen3-4B trained with GRPO, with additional results presented in Appendix~\ref{app:step_action_quality}.  

\begin{figure}[t]
\centering
\includegraphics[width=0.9\textwidth]{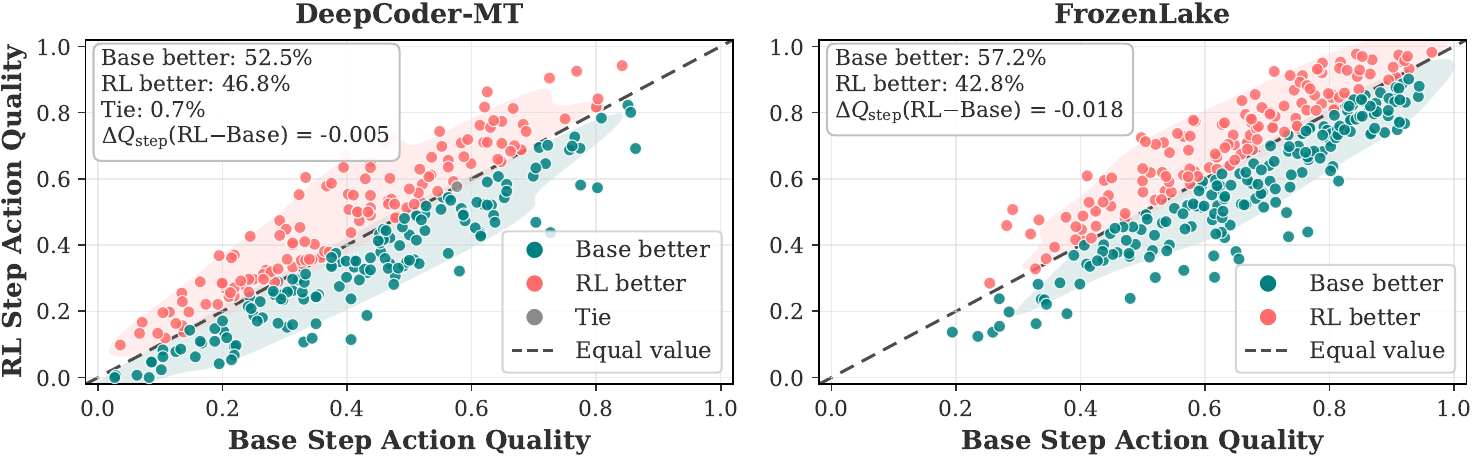}
\caption{Step-level action quality. Each point is one matched state; its coordinates are action quality of the base and RL model. Teal points favor the base model, red points favor the RL model.}
\label{fig:matched}
\vspace{-5pt}
\end{figure}

\textbf{Agentic RL does not uniformly improve step-level action quality.}
We first evaluate whether agentic RL uniformly improves action quality per-step. As shown in Figure~\ref{fig:matched}, each point represents a matched state, with the $x$- and $y$-coordinates indicating action quality of the base and RL models, respectively. Points below the diagonal indicate superior action quality from base model, whereas points above the diagonal favor RL model. We find that the distribution does not favor the RL model and even leans slightly toward the base model. Specifically, RL achieves better step-level decision quality on only 46.8\% of DeepCoder-MT states and 42.8\% of FrozenLake states, with average quality no better than, even slightly below, the base model. These results demonstrate that multi-turn agentic RL does not consistently enhance single-step action and may occasionally degrade it.

\begin{figure}[t]
\centering
\includegraphics[width=0.9\textwidth]{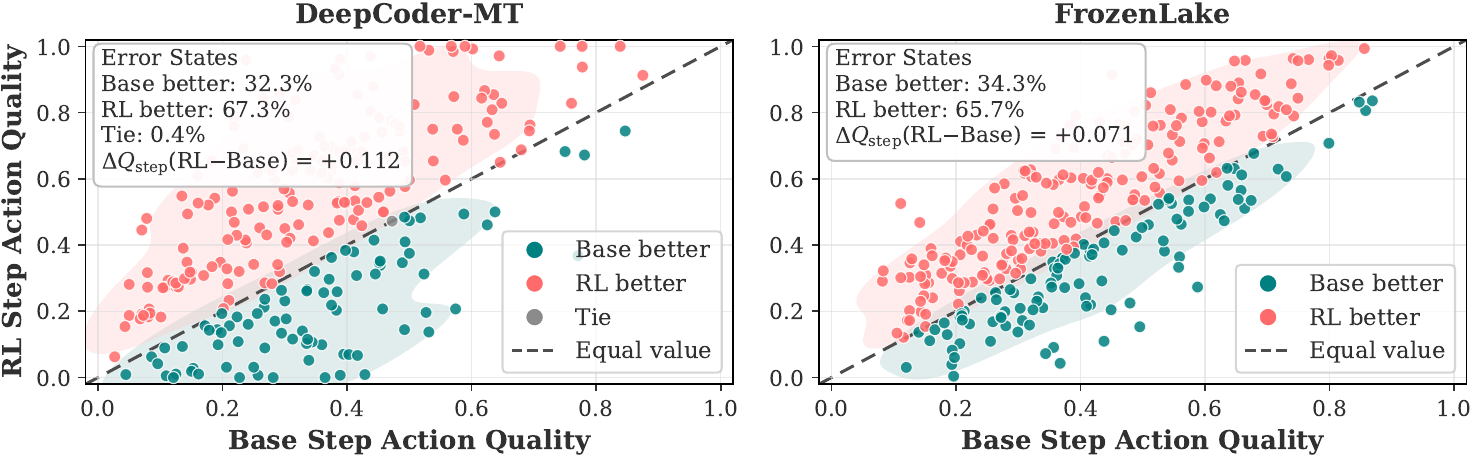}
\caption{Step-level action quality on error states. Each point is one error state.}
\label{fig:error}
\vspace{-13pt}
\end{figure}

\textbf{Agentic RL excels after errors.}
Since the RL model does not consistently yield superior single-step decisions across every matched state, a natural question arises: why does multi-turn RL achieve marked trajectory-level performance gains? To answer this, we focus on states immediately following an error (e.g., error states), where effective recovery ability of models becomes critical. Specifically, we construct an error state pool by injecting an incorrect action at the preceding step. As shown in Figure~\ref{fig:error}, the pattern reverses under these error states: data points shift predominantly above the diagonal, with the RL model selecting the superior action in 65.7\% of cases on FrozenLake and 67.3\% on DeepCoder-MT. We additionally compare the terminal success rates of RL and base models when continuing from identical error states. RL outperforms base on both FrozenLake ($51.3\%$ vs.\ $38.6\%$) and DeepCoder-MT ($38.5\%$ vs.\ $22.9\%$). Appendix~\ref{app:additional_case_study} also illustrates how the RL model corrects an erroneous tool call on BFCL. These results demonstrate that RL achieves higher action quality and continuation success on these error states. When errors occur in interaction trajectories, agentic RL demonstrates superior error recovery capabilities and resilience. 




\begin{figure}[t]
\centering
\includegraphics[width=0.9\textwidth]{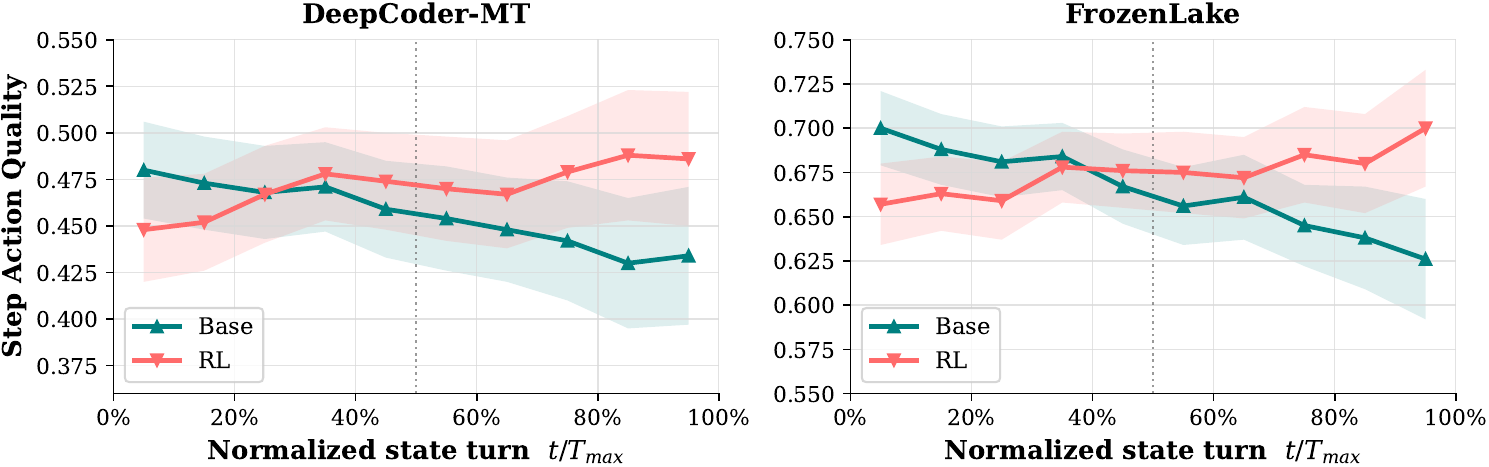}
\caption{Action quality versus normalized trajectory depth
$d_t = t/T_{\max}$. Shading denotes 95\% CI.}
\label{fig:depth}
\end{figure}

\textbf{The RL advantage emerges later in trajectories.}
Finally, we examine how decision quality evolves over the course of a trajectory. We group intermediate states by turn index and define the normalized depth at turn $t$ as $d_t=t/T_{\max}\in[0,1]$, where $T_{\max}$ is the maximum interaction budget. As shown in Figure~\ref{fig:depth}, the RL model shows little advantage early in execution and even underperforms the base model. As interaction proceeds, however, the gap gradually narrows and reverses. This indicates that the RL model does not make superior decisions at the beginning of a trajectory; its advantage emerges as interaction unfolds. This suggests that the step-level advantage of agentic RL is not uniform across a trajectory, but becomes more pronounced at later stages of long-horizon interaction.


\vspace{-2pt}
\begin{tcolorbox}[takeaway,title={Takeaway 3.3}]
Multi-turn agentic RL \textbf{does not consistently improve single-step action quality}. Its advantage is instead more pronounced in error states and later stages of long-horizon trajectories.\end{tcolorbox}
\vspace{-5pt}

\subsection{Risks: Real-World Deployment of Agents}
\label{sec:risk}


These findings highlight a deployment risk: the gains from agentic RL depend on sufficient interaction budget and may diminish or reverse under tight budgets. This is particularly concerning for tasks involving irreversible actions, such as financial or database operations \citep{doshi2026towards}, where agents must make high-quality and conservative actions at each step. Yet higher trajectory success does not ensure better individual actions. As LLM agents are increasingly deployed in real-world settings, understanding such limitations beyond final task success becomes increasingly important.

\section{Mechanisms Behind the Observed Behaviors}
\label{sec:mechanism}

The previous section characterized the \textbf{behavior of RL models} at trajectory and step level. This section digs into the \textbf{mechanism}: a likelihood analysis (\S\ref{sec:likelihood}), the training dynamics analysis (\S\ref{sec:dynamics}).

\subsection{Likelihood Analysis: Local Surprisal and Trajectory-Level Policy Shift}
\label{sec:likelihood}

The results in \S\ref{sec:step} show that the per-step actions of the agent RL model do not consistently outperform those of the base model. In this section, we compare both the RL and base policies through their likelihoods. Specifically, we evaluate rollouts from both models under the base and RL policies, measuring the log-likelihood of individual actions and the relative likelihood of complete trajectories.
Concretely, let a trajectory be $\tau=(h_1,a_1,\ldots,h_T,a_T)$, where $h_t$ denotes the history and $a_t$ is the corresponding action. We keep the histories fixed and evaluate each action under base policy. As an action is a token sequence $a_t=(w_{t,1},\ldots,w_{t,m_t})$, its \textbf{negative log-likelihood (NLL)} is
  \begin{equation*}
  \mathrm{ActionNLL}_{\mathrm{B}}(a_t \mid h_t)
  = -\sum_{j=1}^{m_t}
  \log p_{\mathrm{B}}
  \bigl(w_{t,j}\mid h_t,w_{t,<j}\bigr).
  \end{equation*}
To examine the complete trajectory, we further compute the cumulative \textbf{log-likelihood ratio (LLR)},
  \begin{equation*}
  \mathrm{LLR}_{1:t}(\tau)
  =\sum_{i=1}^{t}
  \left[
  \log \pi_{\mathrm{RL}}(r_i\mid h_i)
  -\log \pi_{\mathrm{B}}(r_i\mid h_i)
  \right],
  \end{equation*}
where $r_i$ denotes the complete model response in one turn, including reasoning and action tokens. Positive LLR indicates that the recorded trajectory is \textit{more likely under the RL policy than under the base policy}. Together these two quantities characterize, respectively, how unusual each single action of the RL model is under the base policy and relative policy preference over complete trajectory.

\begin{wrapfigure}{r}{0.55\textwidth}
\centering
\includegraphics[width=0.55\textwidth]{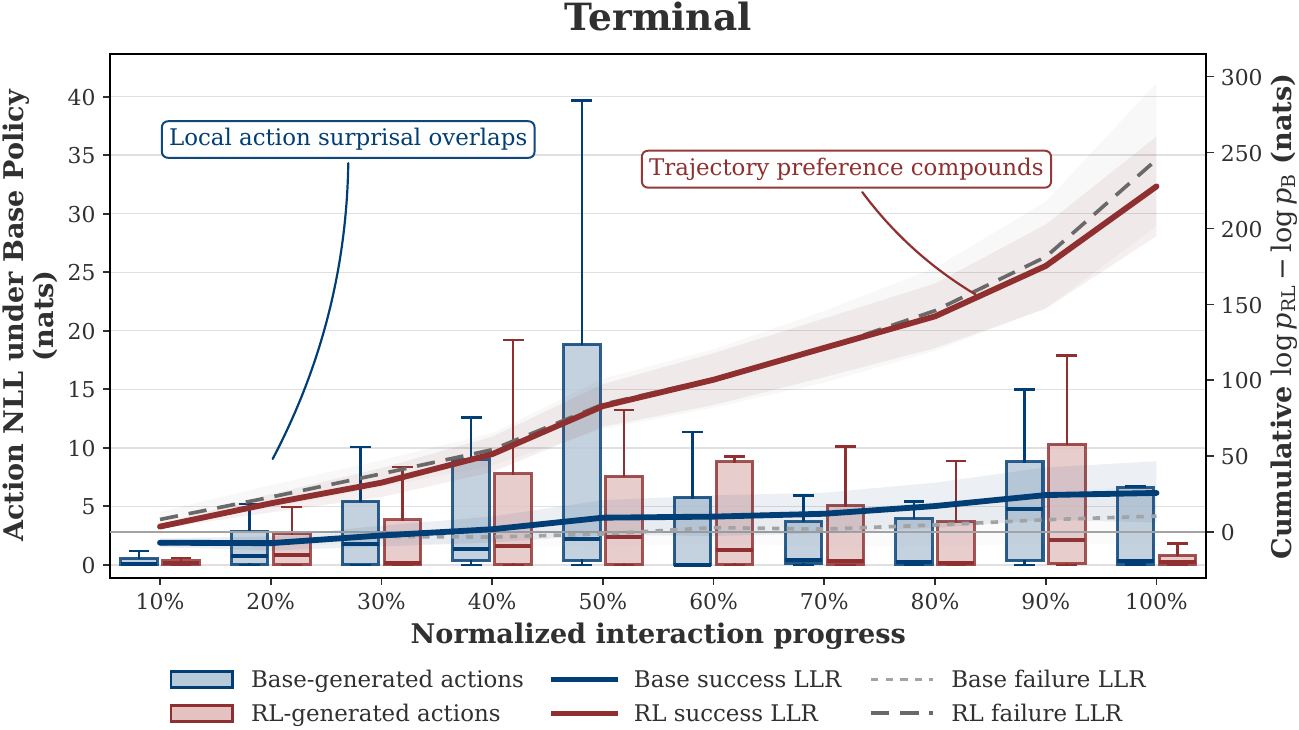}
\vspace{-15pt}
\caption{Likelihood analysis. Boxplots show action NLL under base policy  (left axis). Curves show cumulative LLR (right axis). Bands indicate 95\% confidence.}
\label{fig:nll_analysis}
\vspace{-13pt}
\end{wrapfigure}
  
\textbf{Step-level actions of RL remain locally plausible under base policy.} We evaluate per-token action NLL under the base policy on Terminal-Bench (Figure~\ref{fig:nll_analysis}). We find that the RL action distributions (pink boxplots) substantially overlap with those of the base model (blue boxplots), with mean NLLs of $12.20$ and $10.25$, respectively. Thus, the single-step action of the RL model does not deviate far from the support of the base model, which partly explains why the step-level decisions of the RL model do not consistently beat the base model in \S\ref{sec:step}.

\textbf{From actions to long-horizon execution.}
Although individual actions are supported, this does not mean base model can reliably sample the entire long-horizon trajectory. As shown on the right y-axis of Figure~\ref{fig:nll_analysis}, the cumulative LLR of successful RL trajectories increases with interaction progress, reaching approximately $228$, compared with $26$ for successful base trajectories. The RL model assigns much higher probabilities to its trajectories than base model. This contrasts with the similar NLLs that base model assigns to RL and its own actions. Overall, this demonstrates that even if the conditional probability per-step is reasonably high, the overall probability of entire trajectory can be extremely low. This provides a possible explanation for observations in \S\ref{sec:trajectory}: multi-turn RL enables trajectory-level success to surpass base model's $\mathrm{Pass}@K$ capability boundary even without improvement of every action.

\subsection{Training Dynamics of Multi-Turn RL}
\label{sec:dynamics}

\begin{wrapfigure}{r}{0.55\textwidth}
  \centering
  \includegraphics[width=0.55\textwidth]{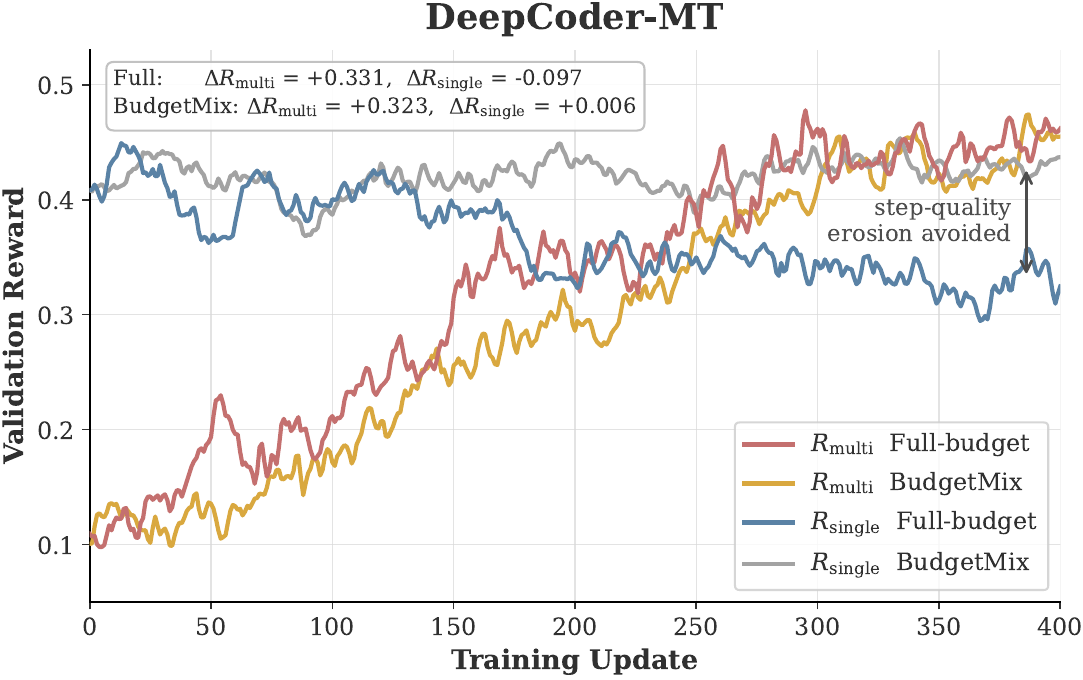}
  \caption{Training dynamics of the trajectory-level reward $R_{\mathrm{multi}}$ and step reward $R_{\mathrm{single}}$ on DeepCoder-MT. }
  \label{fig:training-dynamic}
\vspace{-7pt}
\end{wrapfigure}

To more directly examine how multi-turn RL affects step-level action quality and trajectory-level performance, we further analyze the training dynamics. Standard agentic RL is optimized and validated using trajectory reward $R_{\mathrm{multi}}$. For tasks with step-level quality measures (\S\ref{sec:step}), we track the average step reward $R_{\mathrm{single}}$ as a validation for step-level action, while training optimizes only on $R_{\mathrm{multi}}$. As shown in Figure~\ref{fig:training-dynamic} \textbf{(red and blue curves)}, we find a phenomenon: as training proceeds, the trajectory-level reward $R_{\mathrm{multi}}$ keeps rising while the per-step reward does not improve and even gradually declines. On DeepCoder-MT, $R_{\mathrm{multi}}$ rises by $0.331$ while $R_{\mathrm{single}}$ drops by $0.097$. This indicates that the phenomenon observed at the final checkpoint is not incidental, but rather emerge progressively during training (More results in Appendix~\ref{app:additional_training_dynamics}). This echoes our earlier findings: trajectory-level gains can develop alongside declining step-level action quality.

\begin{figure}[t]
\centering
\includegraphics[width=0.95\textwidth]{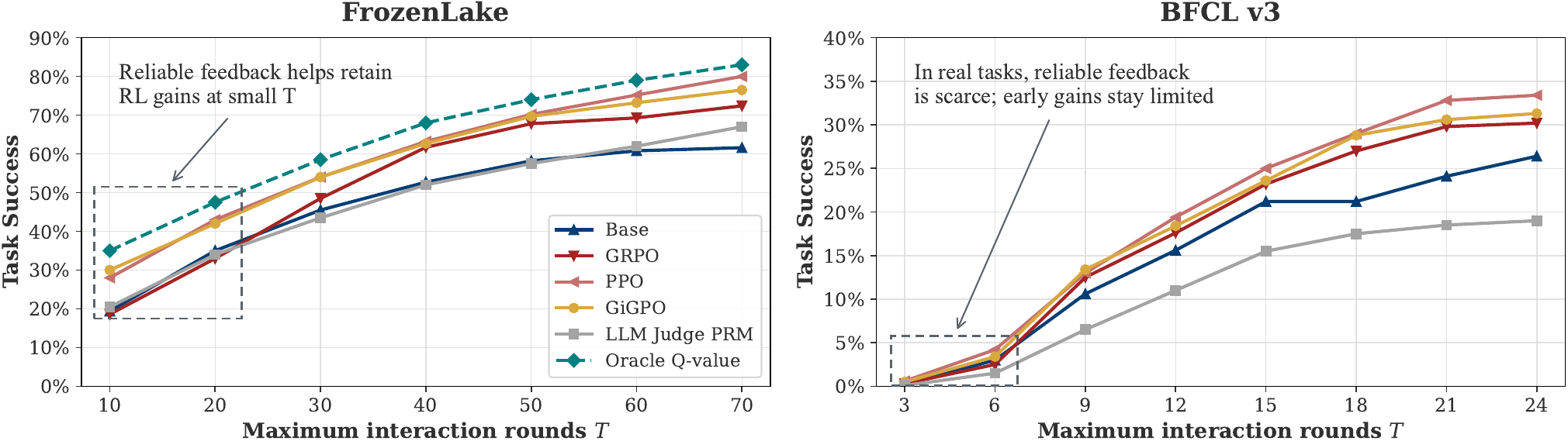}
\caption{Effect of process supervision signals on $\mathrm{Pass}@T$.}
\label{fig:prm-ablation}
\vspace{-7pt}
\end{figure}

\section{Discussion: Towards More Stable Agentic RL}
\label{sec:method}

The preceding sections indicate that the gains of agent RL rely on sufficient interaction budgets, while local action quality does not consistently improve and can even decline. This raises reliability concerns for real-world deployment. Consequently, this \textbf{section offers a preliminary discussion on potential mitigation strategies}. Our objective is to \textrm{preserve its long-horizon execution capabilities and and trajectory-level gains} while reducing its reliance on extra interaction budget for downstream error recovery and encouraging greater prudence in single-step decision making.

\subsection{High-quality Process Signals Can Partly Mitigate this Risk}
\label{sec:effect_prm}

A natural direction is to introduce finer-grained signals to improve per-step action quality through either process rewards or credit assignment. We therefore evaluate several methods to examine whether step-level signals can alleviate above limitations. As shown on the left of Figure~\ref{fig:prm-ablation}, on FrozenLake, accurate process supervision can substantially mitigate the limitations of outcome-only RL. In particular, oracle Q-value process rewards and GiGPO stabilize performance across various interaction budgets, especially under tight budgets. However, as discussed in \S\ref{sec:step}, such precise process supervision is rarely accessible in realistic agent tasks. Results of BFCL on the right illustrate this bottleneck, where no exact step-level oracle is available. Dense signal methods such as GiGPO and PPO fail to enhance performance under constrained budgets and perform similarly to GRPO. These findings demonstrate that while \textbf{high-quality process rewards can alleviate budget reliance, acquiring such supervision signals remains impractical for most real-world applications}.

\subsection{BudgetMix RL: Optimizing RL across Interaction Budgets}
\label{sec:budgetmix}

We therefore take a simpler route: \textbf{BudgetMix RL}. It keeps the same reward mechanism as traditional multi-turn RL, but varies the interaction budget across training trajectories. BudgetMix RL can be combined with existing agentic RL algorithms (e.g., GRPO, PPO, and GiGPO). Standard multi-turn agentic RL samples trajectories under a fixed maximum budget $T_{\max}$ and optimizes either outcome or per-step rewards. However, this objective does not explicitly distinguish solutions that require different amounts of interaction, provided they finish within the budget. \textbf{BudgetMix} makes a single modification to this procedure. For a task instance $x$ with full budget $T_{\max}(x)$, we construct candidate budgets at $10\%$ intervals and uniformly sample one budget per task group:
  \begin{equation}
  \label{eq:budget-assign}
  \mathcal{T}(x)
  =
  \operatorname{Unique}\left(
  \left\{
  \left\lceil \frac{k}{10}T_{\max}(x)\right\rceil
  : k=1,\ldots,10
  \right\}
  \right),
  \qquad
  T(x)\sim\operatorname{Uniform}\bigl(\mathcal{T}(x)\bigr).
  \end{equation}
During training, the total and remaining budget are provided within the prompt; trajectories that exceed the budget without completion receive a reward of zero. The reward signal design is consistent with that of traditional RL. We provide pseudocode and implementation details in Appendix~\ref{app:budgetmix}. BudgetMix turns the budget $T$ from an evaluation dimension in $\mathrm{Pass}@(K,T)$ into a training dimension. Its central idea is that the interaction budget changes the conditions under which the same reward can be obtained. A tighter budget limits opportunities for retries, placing greater demands on earlier decisions, whereas an ample budget preserves opportunities for long-range execution. 

\begin{table*}[t]
\centering
\normalsize
\renewcommand{\arraystretch}{1.08}
\setlength{\tabcolsep}{3.6pt}
\caption{
\(\mathrm{Pass}@T\) evaluation results (\(\mathrm{Pass}@8\), \%).
We evaluate models at 25\%, 50\%, 75\%, and 100\% of the
maximum interaction budget. \textbf{Bold} and \underline{underlined} values denote the best and second-best results, respectively, within each model size and budget. Avg.\ gain reports the average improvement of BudgetMix over full-budget RL across GRPO, GiGPO, and PPO, in percentage points.
}
\label{tab:budgetmix}
\resizebox{\linewidth}{!}{%
\begin{tabular}{lccccccccccccccccc}
\toprule
& \multicolumn{4}{c}{\textbf{FrozenLake} ($T_{\max}=70$)}
& \multicolumn{4}{c}{\textbf{DeepCoder-MT} ($T_{\max}=10$)}
& \multicolumn{4}{c}{\textbf{BFCL-v3} ($T_{\max}=24$)}
& \multicolumn{4}{c}{\textbf{InterCode-SQL} ($T_{\max}=10$)}
& \\
\cmidrule(lr){2-5}
\cmidrule(lr){6-9}
\cmidrule(lr){10-13}
\cmidrule(lr){14-17}
\textbf{Method}
& 25\% & 50\% & 75\% & 100\%
& 25\% & 50\% & 75\% & 100\%
& 25\% & 50\% & 75\% & 100\%
& 25\% & 50\% & 75\% & 100\%
& \textbf{Avg.} \\
\midrule
\multicolumn{18}{c}{\textbf{Qwen3-4B}} \\[-1pt]
\midrule
\rowcolor{BaseGray}\textrm{Base}
& 30.3 & 49.1 & 58.7 & 61.6
& 34.0 & 43.5 & 47.3 & 48.3
& 3.0 & 15.6 & 21.2 & 26.4
& 23.1 & 28.8 & 31.8 & 33.5 & 34.8 \\
\addlinespace[1.2pt]
\textrm{GRPO Full}
& 28.7 & 55.1 & 68.1 & 72.4
& 29.2 & 48.0 & 56.0 & 59.0
& 2.5 & 17.6 & 27.0 & 30.2
& 23.1 & 32.2 & 37.4 & \underline{46.0} & 39.5 \\
\rowcolor{mygray}\textrm{GRPO BudgetMix}
& 37.6 & 61.1 & 72.0 & 73.5
& \textbf{38.1} & \textbf{54.3} & \textbf{60.5} & \textbf{61.2}
& \underline{8.5} & 23.0 & 30.5 & 30.8
& \textbf{28.9} & \textbf{37.7} & \textbf{42.9} & \textbf{47.5} & 44.3 \\
\addlinespace[1.2pt]
\textrm{GiGPO Full}
& 38.4 & 58.4 & 70.4 & 76.5
& 28.8 & 48.3 & 55.6 & 58.8
& 3.4 & 18.4 & 28.8 & 31.3
& 23.4 & 31.5 & 37.5 & 45.0 & 40.9 \\
\rowcolor{mygray}\textrm{GiGPO BudgetMix}
& \textbf{46.3} & \textbf{63.5} & \underline{74.0} & 78.5
& \underline{36.5} & \underline{52.0} & \underline{59.3} & \underline{59.5}
& \underline{8.5} & \underline{23.2} & \underline{31.7} & 33.0
& \underline{27.9} & \underline{37.0} & \underline{41.7} & 44.8 & 44.8 \\
\addlinespace[1.2pt]
\textrm{PPO Full}
& 38.5 & 58.6 & 71.2 & \underline{80.0}
& 29.5 & 45.6 & 51.6 & 55.0
& 4.2 & 19.4 & 29.0 & \underline{33.4}
& 22.4 & 30.6 & 34.8 & 41.0 & 40.3 \\
\rowcolor{mygray}\textrm{PPO BudgetMix}
& \underline{45.5} & \underline{63.1} & \textbf{75.1} & \textbf{80.3}
& 36.4 & 51.6 & 55.5 & 56.9
& \textbf{10.2} & \textbf{24.8} & \textbf{32.8} & \textbf{34.5}
& 27.0 & 35.2 & 38.3 & 42.7 & 44.4 \\
\midrule
\rowcolor{Burgundy!20}\textit{Avg.\ gain}
& $+7.9$ & $+5.2$ & $+3.8$ & $+1.1$
& $+7.8$ & $+5.3$ & $+4.0$ & $+1.6$
& $+5.7$ & $+5.2$ & $+3.4$ & $+1.1$
& $+5.0$ & $+5.2$ & $+4.4$ & $+1.0$ & $+4.2$ \\
\midrule
\multicolumn{18}{c}{\textbf{Qwen3-8B}} \\[-1pt]
\midrule
\rowcolor{BaseGray}\textrm{Base}
& 34.0 & 49.5 & 52.7 & 58.5
& 42.8 & 51.6 & 54.6 & 56.8
& 4.8 & 23.2 & 29.4 & 37.8
& 29.8 & 34.9 & 37.6 & 40.8 & 39.9 \\
\addlinespace[1.2pt]
\textrm{GRPO Full}
& 31.9 & 58.8 & 68.5 & 78.0
& 31.5 & 54.5 & 60.8 & \underline{66.5}
& 3.8 & 27.5 & 39.0 & 45.0
& 27.0 & \underline{40.5} & \underline{45.7} & \underline{49.4} & 45.5 \\
\rowcolor{mygray}\textrm{GRPO BudgetMix}
& 40.8 & \underline{65.1} & \underline{72.9} & 79.6
& \textbf{45.5} & \textbf{59.8} & \textbf{64.6} & \textbf{66.8}
& 10.5 & \underline{33.0} & \underline{43.2} & 47.2
& \textbf{33.1} & \textbf{46.5} & \textbf{50.4} & \textbf{50.4} & 50.6 \\
\addlinespace[1.2pt]
\textrm{GiGPO Full}
& 34.6 & 56.5 & 65.5 & 71.5
& 28.7 & 48.2 & 55.7 & 58.7
& 5.6 & 26.8 & 38.8 & 43.5
& 22.8 & 32.9 & 36.1 & 45.6 & 42.0 \\
\rowcolor{mygray}\textrm{GiGPO BudgetMix}
& \underline{40.9} & 60.4 & 69.1 & 72.3
& \underline{44.9} & \underline{57.8} & 60.1 & 60.7
& \underline{10.8} & 30.8 & 42.8 & 43.2
& \underline{32.6} & 39.4 & 40.2 & 47.7 & 47.1 \\
\addlinespace[1.2pt]
\textrm{PPO Full}
& 40.4 & 63.5 & 71.2 & \underline{85.0}
& 34.5 & 51.5 & 57.0 & 64.8
& 7.0 & 29.5 & 41.5 & \underline{48.0}
& 26.5 & 36.0 & 40.8 & 45.0 & 46.4 \\
\rowcolor{mygray}\textrm{PPO BudgetMix}
& \textbf{47.1} & \textbf{67.4} & \textbf{75.2} & \textbf{86.7}
& 44.0 & 56.6 & \underline{61.1} & 65.9
& \textbf{12.2} & \textbf{34.2} & \textbf{46.0} & \textbf{50.5}
& 32.1 & 40.0 & 44.5 & 46.6 & 50.6 \\
\midrule
\rowcolor{Burgundy!20}\textit{Avg.\ gain}
& $+7.3$ & $+4.7$ & $+4.0$ & $+1.4$
& $+13.2$ & $+6.7$ & $+4.1$ & $+1.1$
& $+5.7$ & $+4.7$ & $+4.2$ & $+1.5$
& $+7.2$ & $+5.5$ & $+4.2$ & $+1.6$ & $+4.8$ \\
\bottomrule
\end{tabular}%
}
\vspace{-5pt}
\end{table*}

\subsection{Experimental Setup}
\label{sec:setup5}

\textbf{Tasks and models.} Our main experiments cover four tasks: FrozenLake, DeepCoder-MT, BFCL and InterCode-SQL. We primarily use Qwen3-4B and Qwen3-8B as the base models. 

\textbf{Baselines.}
From the same base checkpoint, we compare standard full-budget RL against BudgetMix RL under identical algorithm settings including GRPO, GiGPO, PPO. 

\textbf{Evaluation protocol.}
We evaluate $\mathrm{Pass}@T$ with the maximum interaction budget set to 25\%, 50\%, 75\%, and 100\% of $T_{\max}$. Consistent with the previous $\mathrm{Pass}@T$ evaluation setup, we inform the model of the total budget limit and the remaining budget within the prompt, as shown in Appendix \ref{app:evaluation_details}.

\subsection{Results and Analysis}
\label{sec:results5}

Table~\ref{tab:budgetmix} reports the main results. First, under tight budget constraints ($25\%$, $50\%$ and $75\%$ $T_{\mathrm{max}}$), BudgetMix outperforms both the base model and standard full-budget RL across all four tasks, achieving average performance gains of 7.5, 5.3 and 4.0 percentage points over full-budget RL, respectively. Second, at full interaction capacity ($100\%$ $T_{\mathrm{max}}$), BudgetMix maintains parity with full-budget RL, demonstrating that its low-budget enhancements do not compromise long-horizon execution capability. Overall, BudgetMix improves the stability of agent performance, delivering strong performance across diverse interaction budgets and retaining strong full-budget results. 

We further examine the training dynamics of BudgetMix. As shown in Figure~\ref{fig:training-dynamic}, during BudgetMix training (yellow and gray curves), the trajectory-level reward $R_{\mathrm{multi}}$ steadily increases, while the step-level reward $R_{\mathrm{single}}$ remains stable or improves slightly. In contrast, standard full-budget agent RL (blue curve) degrades step-level quality. Thus, BudgetMix largely mitigates local decision-quality degradation while preserving long-horizon gains. Consistent results from direct step-level action-quality evaluation are provided in Appendix~\ref{app:step_action_quality}. 

\section{Related Work}
\label{sec:related}

\subsection{Multi-Turn Agentic Reinforcement Learning}

LLM agents are widely used for interactive tasks like coding and tool use \citep{webarena, toolrl, wang2026think, hao2026pushing, men2025agent}, with RL becoming a major training paradigm \citep{jin2026omni, zhu2026beyond}. Many methods extend single-turn outcome-based RL \citep{dapo} to interactive tasks using trajectory-level rewards \citep{agentgymrl,ragen}, enabling applications such as deep research \citep{searchr1,zerosearch}, tool-augmented reasoning \citep{retool}, and web interaction \citep{webrl}. However, sparse terminal rewards make credit assignment increasingly difficult as interaction horizons grow \citep{trace, hindsight}. To address this, methods derive process signals from outcomes \citep{liu2025online, pgpo}, including state-grouped advantages in GiGPO \citep{gigpo}, stepwise reward decomposition in SPA-RL \citep{sparl}, and advantage attribution with entropy-guided exploration in ARPO \citep{arpo}. Other work explicitly models intermediate decision quality \citep{agentprm, toolprm}. Despite these advances, evaluation remains largely centered on trajectory-level success \citep{li2026benchmark, men2026physics, jin2026look}. We therefore analyze multi-turn RL beyond final task success to better understand its gains and limitations.

\subsection{Understanding and Rethinking RL Post-Training}
\label{sec:related-rethink}

Recently, post-training has become a key stage to improving reasoning and planning capabilities of LLMs \citep{guo2025deepseek}, prompting studies of its mechanisms and limitations \citep{sftconflicts,justrl,fu2026rethinking}. For single-turn reasoning, Pass@$K$ analyses suggest that RLVR primarily reweights solutions already reachable by the base model \citep{limitrlvr}. Later work reports boundary expansion with prolonged RL \citep{prorl}, while others attribute apparent gains to recombining existing skills \citep{fxgxfgxllms} or amplifying latent base-model behaviors \citep{invisibleleash}. These findings suggest that final success may obscure the source of RL gains. Other studies examine entropy dynamics \citep{entropymechanism,entropycollapse}, identify high-entropy tokens as key carriers of learning signals \citep{highentropyminority}. Related work further compares SFT and RL in generalization and capability acquisition \citep{sftmemorizes,sftconflicts}. Mechanistic analyses have recently extended to multi-turn agents \citep{ragen-2,Agent_Environment_Survey,hao2026multisteptoolusereinforcementlearning}. Most relevantly, prior work introduced Pass@$(K,T)$ to characterize multi-turn capability boundaries and showed that agentic RL can surpass capability boundary of base model \citep{doesrlexpandcapability}. We go beyond this perspective by analyzing interaction-budget dependence, step quality and mechanisms to identify the sources of RL gains and potential limitations.

\section{Conclusion}

We revisit the benefits of multi-turn agentic RL beyond final task success. Through $\mathrm{Pass}@(K,T)$ evaluation, we demonstrate that RL can extend the empirical capabilities of the base model, though this advantage may diminish or reverse as interaction budgets become more constrained. Our analysis suggests that these gains stem primarily from long-horizon execution and error recovery, rather than consistently superior single-step actions. As this may impact the reliability of agents deployed in real-world settings, we further discuss potential strategies to mitigate this risk and their limitations.

\section*{AI Use Statement}

During the writing of this paper, we used LLMs to assist with grammar checking, text polishing, and improving overall readability. All AI-assisted revisions were carefully reviewed by the authors to ensure that they did not alter the original meaning, technical content, or conclusions. We also used AI assistants to improve the presentation of selected tables and figures. In addition, we evaluated an LLM-based judge as a process reward model in one of our analysis experiments. We ensure that the AI does not alter the results. All experimental data and numerical results were verified by the authors.

\section*{Ethics Statement}

This work studies the capabilities and limitations of multi-turn reinforcement learning for LLM agents, with particular attention to their reliability under different interaction constraints. Our experiments are conducted on benchmark environments covering tool use, interactive coding, planning, and terminal operations. We do not conduct human-subject experiments in this study. Our analysis highlights a potential deployment risk: the gains of agentic RL can shrink under tight interaction budgets, while single-step action quality of RL models does not consistently improve. This limitation can be particularly relevant in applications involving strict interaction budgets or irreversible actions. We therefore believe that agentic systems should be evaluated not only by final task success, but also by their robustness across execution constraints and the reliability of individual decisions.

\section*{Reproducibility Statement}

We provide the experimental settings and implementation details required to reproduce our main results in both the main text and the appendix. The main text describes the evaluation metrics, experimental protocols, and the BudgetMix RL framework. More complete reproducibility details are provided in the appendix. Specifically, Appendix~\ref{app:training_details} describes the details of tasks and training implementation. Appendix~\ref{app:evaluation_details} provides the detailed evaluation protocols. Additional experiments across RL algorithms, model architectures, tasks, and analysis settings are reported in Appendices~\ref{app:additional_pass_k},\ref{app:additional_pass_t} and \ref{app:additional_analyses}. Together, these details are intended to facilitate reproducibility of our results.

\bibliographystyle{iclr2027_conference}
{\hypersetup{urlcolor=black}%
\bibliography{iclr2027_conference}}

\appendix

\section{Training Details}
\label{app:training_details}
\subsection{Tasks and Training Data}
\label{app:tasks_training_data}

We train and evaluate agents on five representative tasks, covering tool calling, interactive code generation, embodied planning, terminal operations.
For BFCL~\citep{bfcl}, we use executable tool environments and interaction trajectories from EnvFactory~\citep{envfactory} as training data, with downstream evaluation restricted to BFCL.
For DeepCoder-MT, we construct multi-turn code generation tasks from DeepCoder~\citep{deepcoder2025} programming problems: at each turn, the model can submit a candidate program, inspect unit-test feedback, and revise the program in subsequent turns.
For FrozenLake, we independently generate training and test maps, ensuring that the two splits are disjoint.
For Terminal-Bench 1.0~\citep{terminalbench}, we use the training tasks provided by Tmax~\citep{tmax}.
For InterCode-SQL~\citep{intercode}, we construct interactive SQL training environments from the Spider training set~\citep{spider} and evaluate on a separate split.

\subsection{Models and Training Hyperparameters}
\label{app:models_training}

We consider Qwen3-4B and Qwen3-8B~\citep{qwen3} and three RL algorithms: GRPO~\citep{guo2025deepseek}, GiGPO~\citep{gigpo}, and PPO~\citep{ppo}. GRPO and GiGPO do not use value networks, whereas PPO trains a separate critic to estimate state values. GiGPO derives step-level credit assignment from terminal rewards and recurring environment states. Neither GiGPO nor PPO requires an additional external process reward model. Our implementation builds on the VeRL framework~\citep{verl}, and we thank its developers for their excellent work.

All algorithms are trained from the base model. The learning rate is $1\times 10^{-6}$, with weight decay of $0.01$. Training rollouts are sampled with temperature $1.0$ and top-$p$ of $1.0$. KL regularization against the reference policy uses the \texttt{low\_var\_kl} estimator with a coefficient of $0.001$. For GRPO and GiGPO, we sample $G=16$ trajectories per task instance. PPO also generates $16$ trajectories per task instance to match the sampling budget, but does not use group advantages. Full-Budget and BudgetMix RL use identical training settings; their only difference is the distribution of budgets.

\textbf{GRPO.}
We compute group-relative advantages over $G$ trajectories sampled for the same task instance. Environment-observation tokens and padding tokens are excluded from the policy loss.

\textbf{GiGPO.}
Following the official GiGPO~\citep{gigpo} settings, we combine trajectory-level advantages with anchor-state step-level advantages:
\begin{equation*}
    A_{i,t}=A_i^E+\omega A_{i,t}^S.
\end{equation*}
In the main experiments, we set the step-level weight to $\omega=1$ and the environment-step discount factor to $\gamma=0.95$.
For environment step $t$ of trajectory $i$, the discounted return is
\begin{equation*}
    R_{i,t}
    =
    \sum_{u=t}^{T_i}\gamma^{u-t}r_{i,u},
\end{equation*}
where $T_i$ is the number of environment steps in trajectory $i$.

\textbf{PPO.}
The critic is initialized from the same base model as the actor, while optimizers remain independent.
The critic learning rate is $1\times 10^{-5}$, with weight decay of $0.01$. The critic is updated from the start of training, without warmup. Both actor and critic are updated for one epoch per rollout batch. We use token-level generalized advantage estimation (GAE)~\citep{gae}. The lower and upper policy-clipping parameters are both set to $0.2$. Value clipping is set to $0.5$.

\subsection{Interaction Budgets}
\label{app:interaction_budgets}

For the $\mathrm{Pass}@T$ evaluations, we set
$T_{\max}=70$ for FrozenLake,
$T_{\max}=10$ for DeepCoder-MT,
$T_{\max}=24$ for BFCL-v3,
$T_{\max}=10$ for InterCode-SQL,
and $T_{\max}=50$ for Terminal-Bench. The maximum number of interaction turns per trajectory is explicitly communicated to the model through the system prompt shown in Table~\ref{tab:budget-aware-prompt}. Here, \colorbox{Burgundy!8}{\textcolor{Burgundy}{\texttt{[ENVIRONMENT]}}}
  denotes the task environment (such as FrozenLake, Terminal-Bench);
  \colorbox{Burgundy!8}{\textcolor{Burgundy}{\texttt{[MAX\_TURNS]}}}
  is the maximum number of permitted interactions;
  \colorbox{Burgundy!8}{\textcolor{Burgundy}{\texttt{[CURRENT\_TURN]}}}
  is the number of completed interactions; and
  \colorbox{Burgundy!8}{\textcolor{Burgundy}{\texttt{[REMAINING\_TURNS]}}}
  is the remaining interaction budget.
  \colorbox{black!5}{\textcolor{Burgundy}{\texttt{[TASK\_SPECIFIC\_PROMPT]}}}
  contains the task instruction, while
  \colorbox{black!5}{\textcolor{Burgundy}{\texttt{[TASK\_SPECIFIC\_ACTION\_FORMAT]}}}
  specifies the required action syntax.

 \begin{table*}[t]
  \centering
  \caption{
  Budget-aware prompt template used during training and evaluation.
  The turn counters are instantiated before each model action. 
  }
  \label{tab:budget-aware-prompt}

  \renewcommand{\arraystretch}{1.25}
  \setlength{\tabcolsep}{10pt}

  \begin{tabular}{>{\raggedright\arraybackslash}p{0.94\textwidth}}
  \toprule

  \cellcolor{Burgundy!10}
  \textcolor{Burgundy}{\textbf{SYSTEM PROMPT}}
  \\[3pt]

  \cellcolor{SoftGray}
  You are an agent interacting with the
  \textcolor{Burgundy}{\texttt{[ENVIRONMENT]}}
  task under a finite interaction budget.
  One interaction turn consists of one action from you followed by one
  observation from the environment.

  \medskip

  \textbf{Maximum number of interaction turns:}
  \textcolor{Burgundy}{\textbf{\texttt{[MAX\_TURNS]}}}.

  \textbf{Number of turns already completed:}
  \textcolor{Burgundy}{\textbf{\texttt{[CURRENT\_TURN]}}}.

  \textbf{Number of turns remaining:}
  \textcolor{Burgundy}{\textbf{\texttt{[REMAINING\_TURNS]}}}.

  \medskip

  Follow the task-specific instruction below.
  At each turn, output exactly one action in the required format.
  Do not output multiple actions in one turn, and do not assume that
  additional turns will be available after the budget is exhausted.
  \\[6pt]

  \cellcolor{Burgundy!7}
  \textcolor{Burgundy}{\textbf{TASK-SPECIFIC INSTRUCTION}}
  \\[3pt]

  \cellcolor{SoftGray}
  \texttt{[TASK\_SPECIFIC\_PROMPT]}
  \\[6pt]

  \cellcolor{Burgundy!7}
  \textcolor{Burgundy}{\textbf{ACTION FORMAT}}
  \\[3pt]

  \cellcolor{SoftGray}
  Output exactly one action in the following format:

  \medskip

  \hfill
  \textcolor{Burgundy}{\texttt{[TASK\_SPECIFIC\_ACTION\_FORMAT]}}
  \hfill\null
  \\

  \bottomrule
  \end{tabular}
  \end{table*}

\subsection{BudgetMix Implementation}
  \label{app:budgetmix}

  \begin{algorithm}[t]
  \caption{BudgetMix with GRPO}
  \label{alg:budgetmix-grpo}
  \begin{algorithmic}[1]
  \Require Policy $\pi_\theta$, dataset $\mathcal{D}$, group size $G$,
  candidate budgets $\mathcal{B}(x)$
  \For{each training iteration}
      \State $\pi_{\mathrm{old}}\gets\pi_\theta$
      \State Sample a task batch $\mathcal{X}$ from $\mathcal{D}$
      \For{each task $x\in\mathcal{X}$}
          \State Sample $B\sim\operatorname{Uniform}(\mathcal{B}(x))$
          \State Insert $B$ into the budget-aware prompt
          \State Collect $G$ trajectories using $\pi_{\mathrm{old}}$,
              each with at most $B$ interaction turns
          \State Assign outcome rewards $R_1,\ldots,R_G$
          \State $A_i\gets
              \dfrac{R_i-\operatorname{mean}(R)}
              {\operatorname{std}(R)+\epsilon}$
              for $i=1,\ldots,G$
      \EndFor
      \State Update $\pi_\theta$ using the standard GRPO objective
          with reference KL regularization
  \EndFor
  \end{algorithmic}
  \end{algorithm}

\textbf{Budget sampling.} BudgetMix follows the training settings of the corresponding Full-Budget RL baseline, changing only the interaction-budget distribution. For each task $x$, we construct candidate budgets at $10\%$ intervals of its full budget $B_{\max}(x)$:
  \begin{equation}
      \mathcal{B}(x)
      =
      \operatorname{Unique}\left(
      \left\{
          \left\lceil \frac{k}{10}B_{\max}(x)\right\rceil
          : k=1,\ldots,10
      \right\}
      \right),
      \qquad
      B\sim\operatorname{Uniform}(\mathcal{B}(x)).
  \end{equation}
The distribution remains fixed throughout training. Algorithm~\ref{alg:budgetmix-grpo} presents the implementation of BudgetMix using GRPO as an example. For GRPO and GiGPO, all trajectories within a task group share one sampled budget. For PPO, the actor and critic receive the same budget conditioning. The prompt specifies the assigned budget and the number of completed interactions. 

\textbf{Rewards and termination.} Trajectories that exhaust their budget without success receive a \textbf{terminal success reward of zero}. For PPO, task termination and budget exhaustion both use zero terminal bootstrap, with trajectories retained for policy and value updates. For GiGPO, we match anchors within the same task instance and rollout group using both the environment state and the remaining interaction budget. Specifically, the anchor is $(s_t,B-t)$, where $t$ counts completed interactions before the current action. Singleton or zero-variance anchor groups receive zero step-level advantage, while the trajectory-level advantage is retained.


\section{Evaluation Details}
\label{app:evaluation_details}

In this section, we provide a detailed description of the evaluation settings and implementation details presented in our paper.

\subsection{Generation Settings and Paired Comparisons}
\label{app:generation_settings}

For all tasks, we evaluate with temperature $0.6$ and top-$p=0.9$.
Base and RL models are evaluated on identical task instances, with the same prompt templates, interaction budgets, and numbers of sampled trajectories.
The budget is specified in the system prompt before generation begins, as illustrated in Table~\ref{tab:budget-aware-prompt}.
Task success within the specified budget is determined by the official evaluator.

\subsection{$\mathrm{Pass}@K$}
\label{app:pass_k_metrics}

\textbf{$\mathrm{Pass}@K$.} For each task instance, we generate $n$ independent trajectories and denote the number of successful trajectories by $c$.
For $n\geq k$, we use the unbiased estimator
\begin{equation*}
    \widehat{\mathrm{Pass}@k}
    =
    1-\frac{\binom{n-c}{k}}{\binom{n}{k}}.
    \label{eq:pass_k_estimator}
\end{equation*}
This estimator equals the probability of obtaining at least one success when uniformly selecting $k$ trajectories without replacement from the $n$ generated trajectories. It reduces variance relative to direct estimation using repeated groups of $k$ samples. In the main text, we sweep $k$ up to $128$. This estimation follows \citet{chen2021evaluating} and \citet{limitrlvr}.

\subsection{$\mathrm{Pass}@T$}
\label{app:pass_T_metrics}

\textbf{$\mathrm{Pass}@T$.} For interactive tasks, we additionally report $\mathrm{Pass}@T$. Each trajectory allows at most $T$ interactions, and performance is measured by the probability that at least one of $k$ independent trajectories succeeds. In the main-text $\mathrm{Pass}@T$ evaluations, we fix $k=8$ and sweep $T$. 

\begin{figure}[t]
    \centering
    \includegraphics[width=\textwidth]{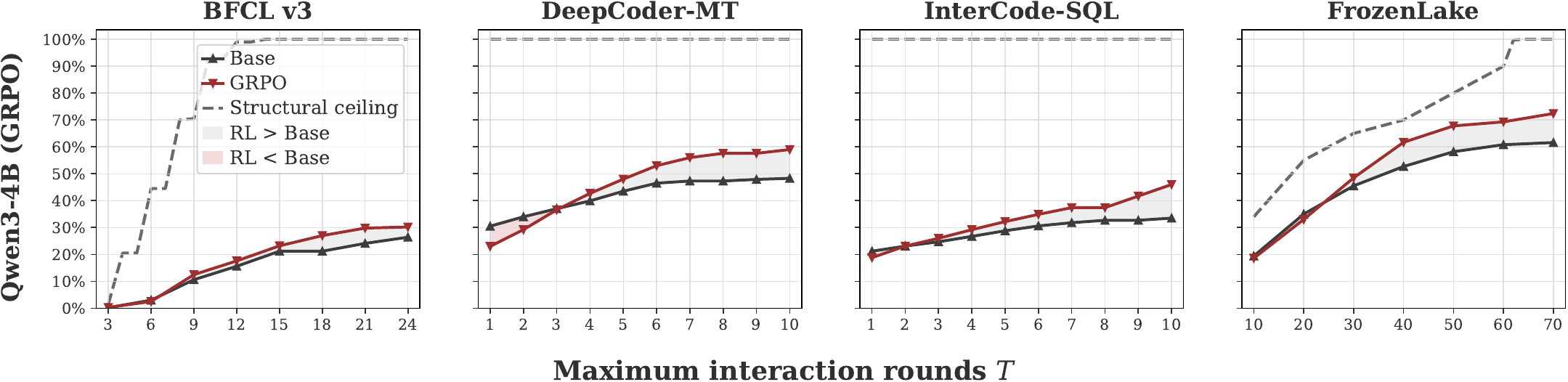}
    \caption{
        Model performance and structural performance upper bounds
        across interaction budgets.
        Gray lines indicate the structural upper bounds.
    }
    \label{fig:structural_budget_bounds}
\end{figure}

Some tasks may become structurally unsolvable under overly restrictive budgets. To clarify that our findings in the main text are not driven by this constraint, we also plot the structural performance upper bound for each task at each budget. For FrozenLake, we compute the shortest path for each task instance and use it to determine the structural performance upper bound under a given interaction budget. For BFCL, we estimate this upper bound using the officially annotated ground-truth tool-call trajectories. For DeepCoder and InterCode-SQL, each turn consists of a complete code submission, meaning that every instance is, in principle, solvable within a single turn; their structural upper bounds therefore remain at 100\%. As shown in Figure~\ref{fig:structural_budget_bounds}, at small $T$, these upper bounds, shown by the gray lines, decrease on some tasks, notably FrozenLake and BFCL. Nevertheless, substantial headroom remains for both models, yet RL still shows no advantage over base model. This supports our conclusion that the gains of agentic RL depend on sufficiently generous interaction budgets.

\subsection{Local Decision Quality at Matched States}
\label{app:matched_state_evaluation}

As described in \S\ref{sec:step}, we extract intermediate states from complete base and RL trajectories and construct a state pool with equal contributions from the two policies to mitigate bias.
Each state includes the current observation and the history.
For each state, we sample $G=16$ next-step actions from each policy and estimate its single-step decision quality by the sample mean of the corresponding action values.
For FrozenLake, we use the known deterministic transition dynamics of the environment to compute the exact $Q^*$ value of each candidate action via finite-horizon dynamic programming (DP) or breadth-first search (BFS).
For DeepCoder-MT, we execute each candidate program independently and use its pass rate on the official unit tests as the action-quality measure.

\subsection{Error-State Construction}
\label{app:error_state_construction}

We construct error states to evaluate single-step decision quality as follows.

For FrozenLake, we select a random intermediate state $s_t$ along a trajectory and execute a randomly chosen action $a_t$. If the action increases the shortest-path distance to the goal, we regard the resulting state as worse than the preceding state and include it in the error-state pool.

For DeepCoder-MT, we use Qwen3-1.7B to modify the code at a randomly selected intermediate state. If the modified program passes fewer unit tests than the original program, we regard this as a degradation in code quality and use the resulting state for subsequent evaluation.

Both constructions serve as proxies rather than rigorous definitions of error states.

\subsection{Effect of Process Supervision Signals}

In \S\ref{sec:effect_prm} of the main text, we preliminarily investigate whether high-quality process supervision can alleviate the dependence of agentic RL gains on interaction budgets. On FrozenLake, we use oracle $Q$-values as process rewards, following the same procedure as in \S~\ref{sec:step}: we compute the $Q$-value of each action via finite-horizon dynamic programming. For both tasks, we additionally explore using an LLM judge as a process reward model to provide step-level supervision. Specifically, we use GPT-5.6-Sol as the judge model with the prompt shown in Table~\ref{tab:llm-judge-prompt}. Our results show that even a strong judge model cannot consistently provide high-quality process supervision across all tasks, particularly in more realistic settings such as BFCL.

\begin{table*}[t]
    \centering
    \caption{
    LLM judge prompt for step-level process supervision.
    The judge scores the current action before execution.
    Scores are mapped to $[0,1]$ by dividing by 10.
    }
    \label{tab:llm-judge-prompt}

    \renewcommand{\arraystretch}{1.25}
    \setlength{\tabcolsep}{10pt}

    \begin{tabular}{>{\raggedright\arraybackslash}p{0.94\textwidth}}
    \toprule

    \cellcolor{Burgundy!10}
    \textcolor{Burgundy}{\textbf{SYSTEM PROMPT}}
    \\[3pt]

    \cellcolor{SoftGray}
    You are evaluating one action taken by an agent in an
    interactive task.

    \medskip

    Given the task, interaction history, current observation, and remaining interaction budget, rate the proposed action with an integer from 0 to 10. Evaluate its correctness, relevance, and expected contribution to completing the task efficiently. Necessary intermediate steps can receive full credit. Do not reward verbosity or redundant actions.
    \\[6pt]

    \cellcolor{Burgundy!7}
    \textcolor{Burgundy}{\textbf{INPUT}}
    \\[3pt]

    \cellcolor{SoftGray}
    \textbf{Task:}
    \textcolor{Burgundy}{\texttt{[TASK]}}

    \medskip

    \textbf{Interaction history:}
    \textcolor{Burgundy}{\texttt{[HISTORY]}}

    \medskip

    \textbf{Current observation:}
    \textcolor{Burgundy}{\texttt{[OBSERVATION]}}

    \medskip

    \textbf{Remaining turns, including this action:}
    \textcolor{Burgundy}{\texttt{[REMAINING\_TURNS]}}

    \medskip

    \textbf{Action to evaluate:}
    \textcolor{Burgundy}{\texttt{[ACTION]}}
    \\[6pt]

    \cellcolor{Burgundy!7}
    \textcolor{Burgundy}{\textbf{OUTPUT FORMAT}}
    \\[3pt]

    \cellcolor{SoftGray}
    Output only one integer from 0 to 10, with no explanation.
    \\

    \bottomrule
    \end{tabular}
  \end{table*}

\section{Additional \texorpdfstring{$\mathrm{Pass}@K$}{Pass@K} Experiments}
\label{app:additional_pass_k}

Due to space limitations, the main text presents only the $\mathrm{Pass}@K$ results for Qwen3-4B and Qwen3-8B trained with GRPO.
Here, we provide additional results across agentic RL algorithms, base model architectures, and a more realistic task to further substantiate our findings.

\subsection{Additional Agentic RL Algorithms}
\label{app:pass_k_algorithms}

We first evaluate additional agentic RL algorithms to examine whether our $\mathrm{Pass}@K$ findings extend beyond outcome-only GRPO.
We consider two types of methods:
GiGPO~\citep{gigpo}, which uses credit assignment to derive richer, denser learning signals from outcome rewards, and
PPO~\citep{ppo}, which uses a critic to provide denser process-level learning signals.

\begin{figure}[t]
    \centering
    \includegraphics[width=\textwidth]{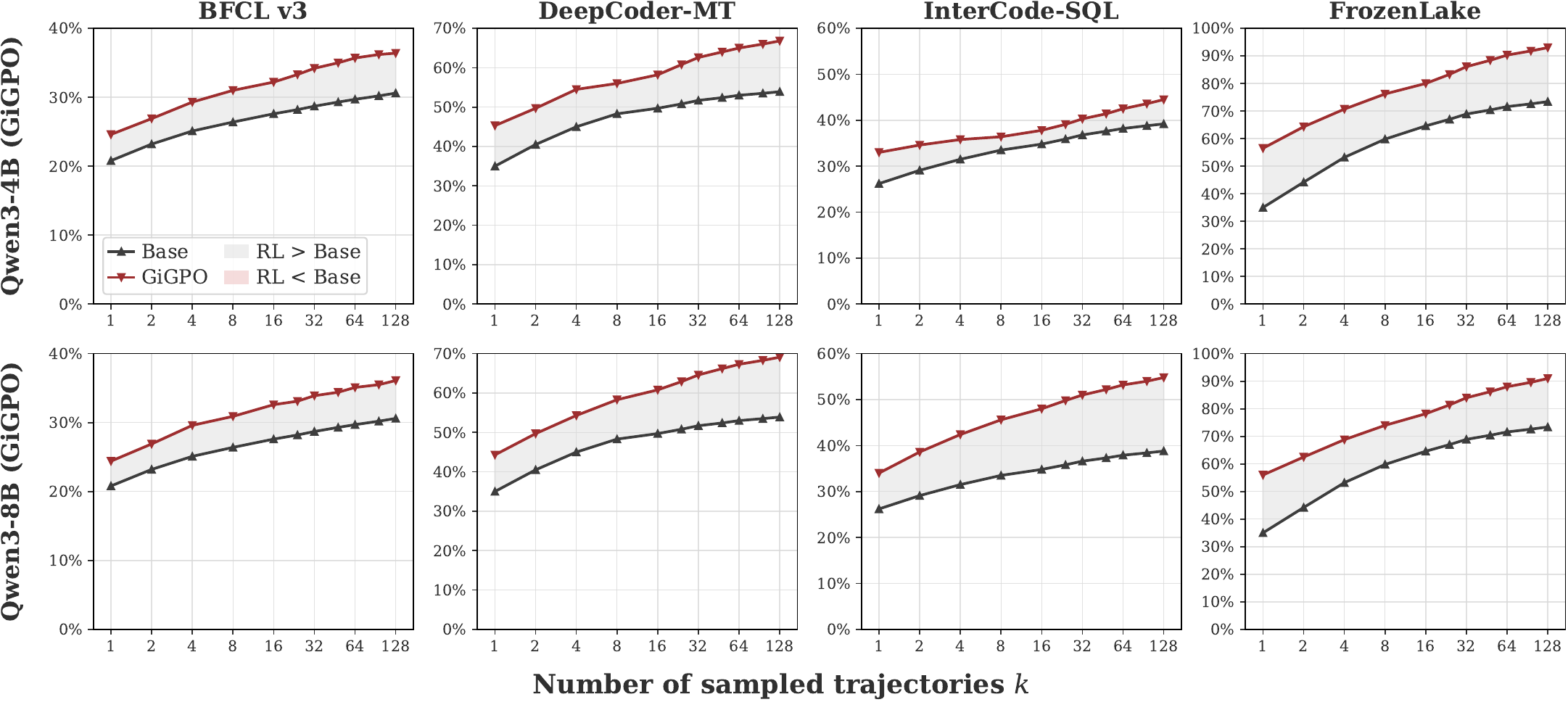}
    \caption{
        Additional $\mathrm{Pass}@K$ results with GiGPO.
    }
    \label{fig:pass_k_algorithms_gigpo}
\end{figure}

\begin{figure}[t]
    \centering
    \includegraphics[width=\textwidth]{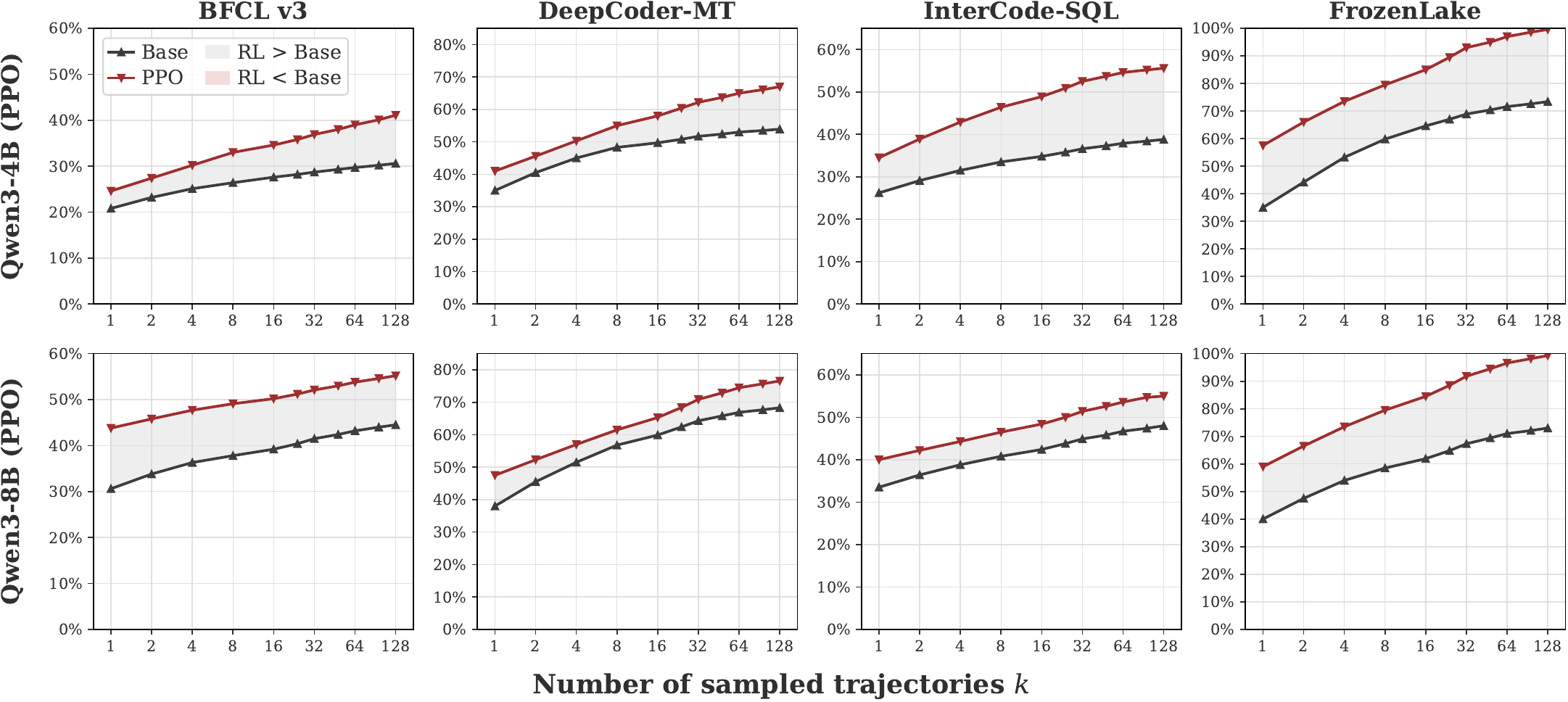}
    \caption{
        Additional $\mathrm{Pass}@K$ results with PPO.
    }
    \label{fig:pass_k_algorithms_ppo}
\end{figure}

As shown in Figure~\ref{fig:pass_k_algorithms_gigpo} and ~\ref{fig:pass_k_algorithms_ppo}, our conclusion holds for both algorithms:
RL-trained models consistently outperform their base models as the number of sampled trajectories $K$ increases.

\subsection{Additional Model Architectures}
\label{app:pass_k_architectures}

\begin{figure}[t]
    \centering
    \includegraphics[width=0.9\textwidth]{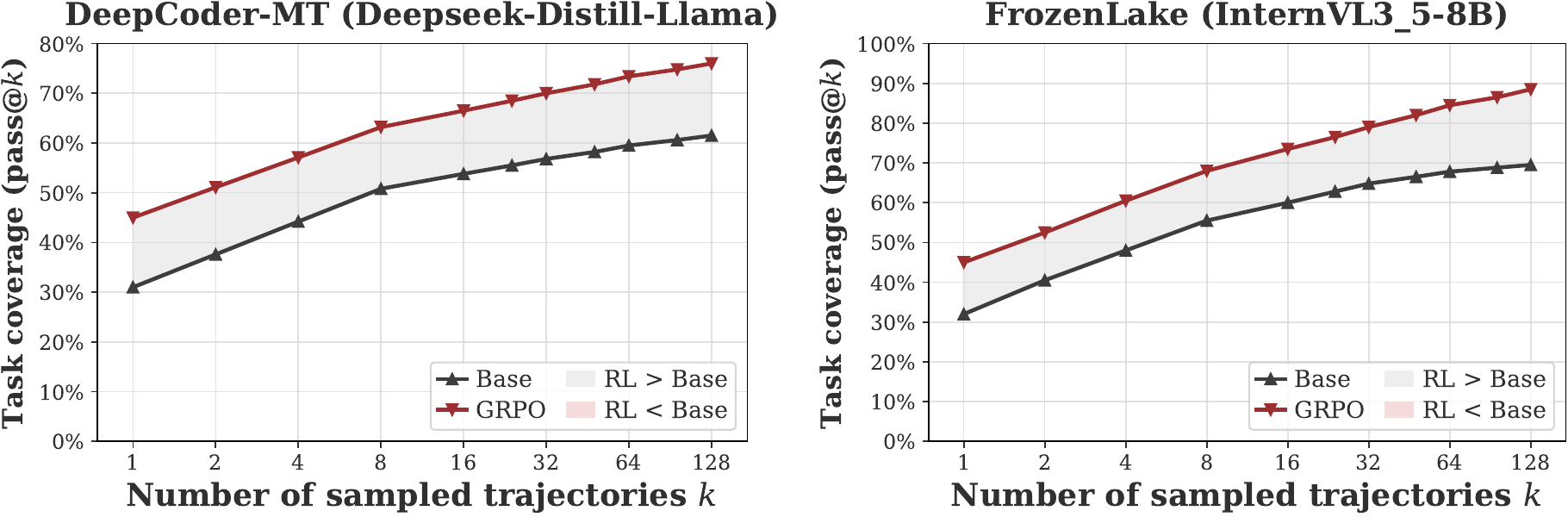}
    \caption{
        $\mathrm{Pass}@K$ results for InternVL3.5-8B on FrozenLake
        and DeepSeek-R1-Distill-Llama-8B on DeepCoder-MT with GRPO.
    }
    \label{fig:pass_k_architectures}
\end{figure}

We evaluate InternVL3.5-8B~\citep{internvl35} on FrozenLake and DeepSeek-R1-Distill-Llama-8B~\citep{guo2025deepseek} on DeepCoder-MT.
For each model, we conduct RL training with GRPO. Figure~\ref{fig:pass_k_architectures} shows that agentic RL consistently surpasses the empirical $\mathrm{Pass}@K$ capability boundaries of base models across these architectures. This finding is therefore not specific to the Qwen3 family.

\subsection{Additional Tasks}
\label{app:pass_k_tasks}

\begin{figure}[t]
    \centering
    \includegraphics[width=0.9\textwidth]{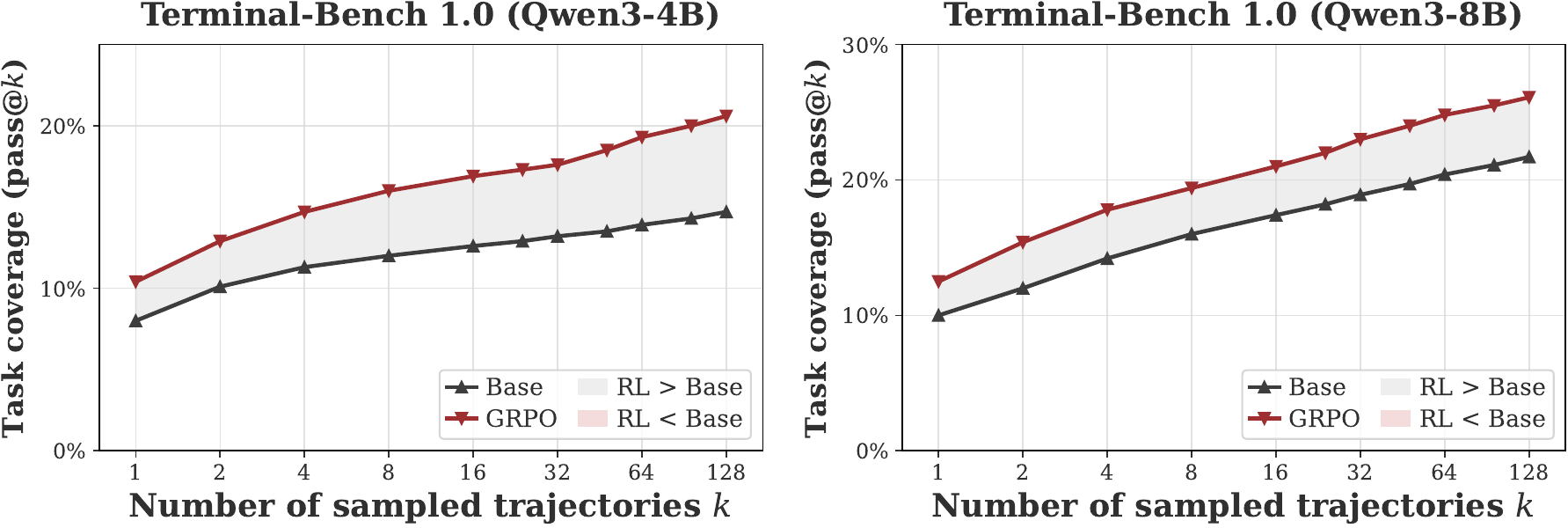}
    \caption{
        Additional $\mathrm{Pass}@K$ results on Terminal-Bench.
    }
    \label{fig:pass_k_terminal}
\end{figure}

Finally, we include Terminal-Bench~\citep{terminalbench}, a more realistic agent benchmark, to examine whether our findings remain robust in more realistic environments. Figure~\ref{fig:pass_k_terminal} shows a pattern consistent with the preceding experiments, supporting the relevance of our findings to realistic settings. For both scales, RL models consistently outperform their base models as $K$ increases. These results further suggest that agentic RL can reliably expand the empirical capability boundary.

\clearpage

\section{Additional \texorpdfstring{$\mathrm{Pass}@T$}{Pass@T} Experiments}
\label{app:additional_pass_t}

As in Appendix~\ref{app:additional_pass_k}, we vary the RL algorithm, model architecture, and task to examine the generality of our $\mathrm{Pass}@T$ findings:
the gains of agentic RL depend on sufficient interaction budgets, and its advantage over the base model shrinks, disappears, or even reverses as the turn budget $T$ tightens.
Does this pattern hold across all settings?

\subsection{Additional Agentic RL Algorithms}
\label{app:pass_t_algorithms}

Whereas the main text primarily reports results with outcome-only GRPO, this section examines GiGPO and PPO.
Both provide process-level learning signals:
GiGPO derives finer-grained credit assignment from outcome rewards, while PPO trains a critic to provide denser learning signals.

\begin{figure}[t]
    \centering
    \includegraphics[width=\textwidth]{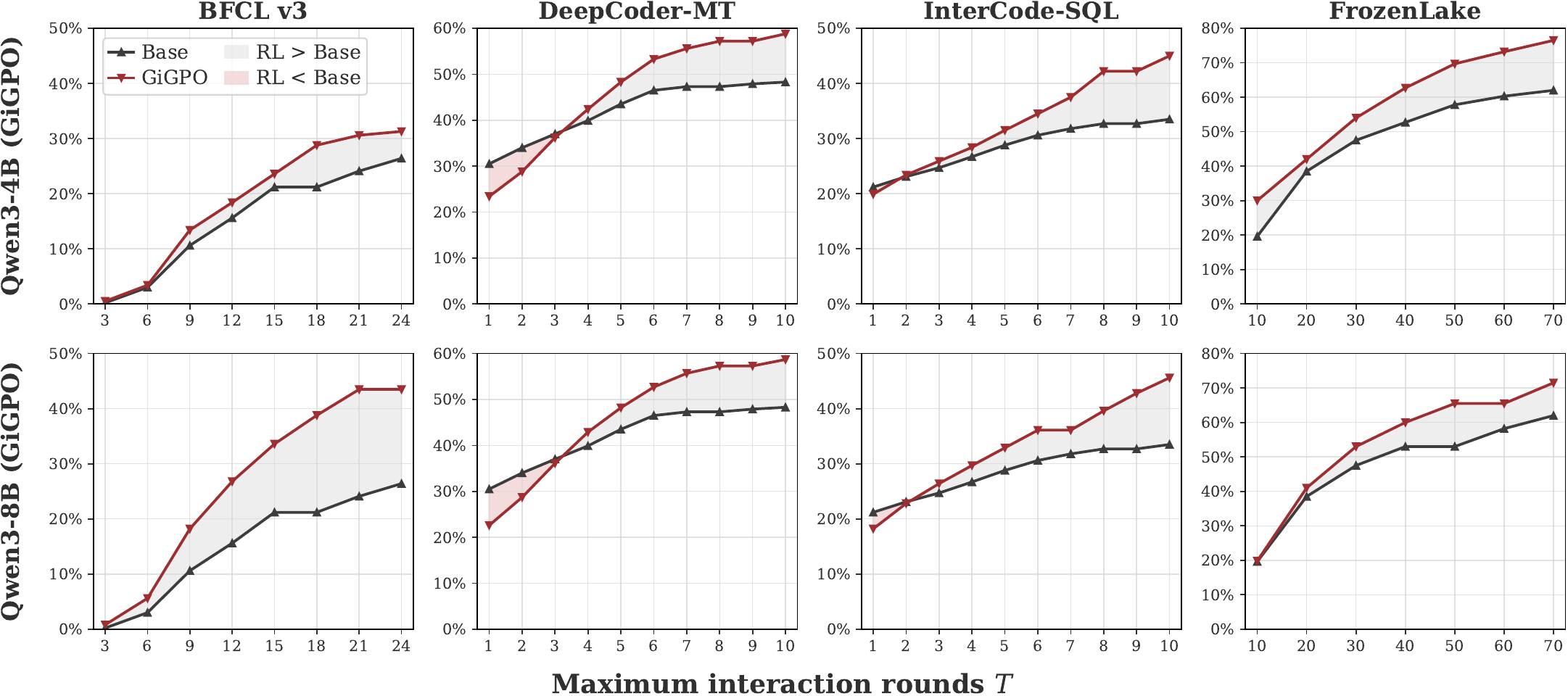}
    \caption{
        $\mathrm{Pass}@T$ results with GiGPO
        on FrozenLake, DeepCoder-MT, BFCL and InterCode-SQL.
    }
    \label{fig:pass_t_algorithms_gigpo}
\end{figure}

\begin{figure}[t]
    \centering
    \includegraphics[width=\textwidth]{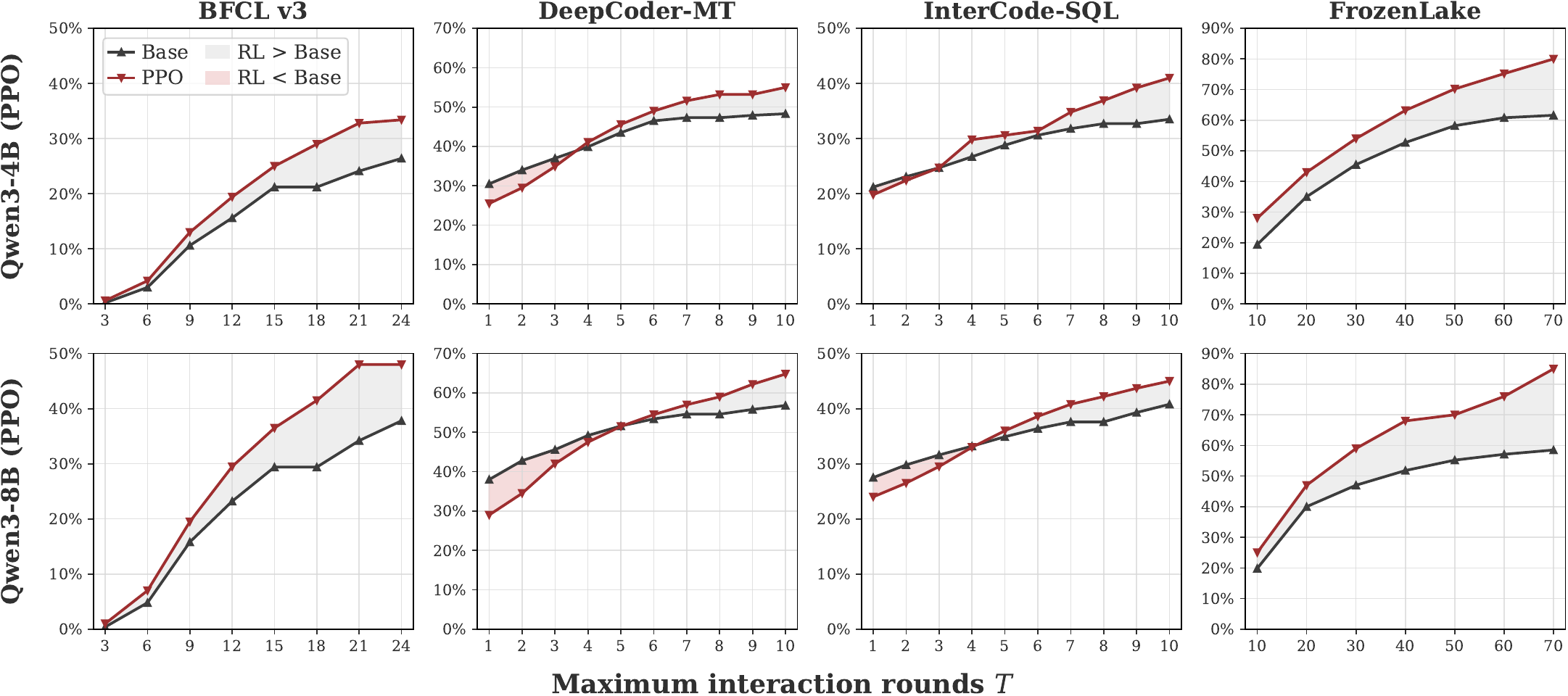}
    \caption{
        $\mathrm{Pass}@T$ results with PPO
        on FrozenLake, DeepCoder-MT, BFCL and InterCode-SQL.
    }
    \label{fig:pass_t_algorithms_ppo}
\end{figure}

Figure~\ref{fig:pass_t_algorithms_gigpo} and ~\ref{fig:pass_t_algorithms_ppo} reveal an interesting contrast.
FrozenLake is a relatively simple, idealized environment in which single-step decision quality and process rewards can be evaluated reliably.
On this task, both GiGPO and PPO alleviate, to varying degrees, the $\mathrm{Pass}@T$ limitation:
rather than losing their advantage as the budget tightens, models trained with these algorithms achieve relatively consistent gains across budgets.
However, on more realistic tasks such as DeepCoder-MT and InterCode-SQL, reliable process supervision is harder to obtain, and neither algorithm effectively alleviates this limitation. This is consistent with our analysis in \S\ref{sec:method}:
high-quality process supervision can improve single-step decision quality and stabilize gains across interaction budgets.
However, such supervision is difficult to obtain in more realistic tasks, which limits its effectiveness.

\subsection{Additional Model Architectures}
\label{app:pass_t_architectures}

\begin{figure}[t]
    \centering
    \includegraphics[width=0.9\textwidth]{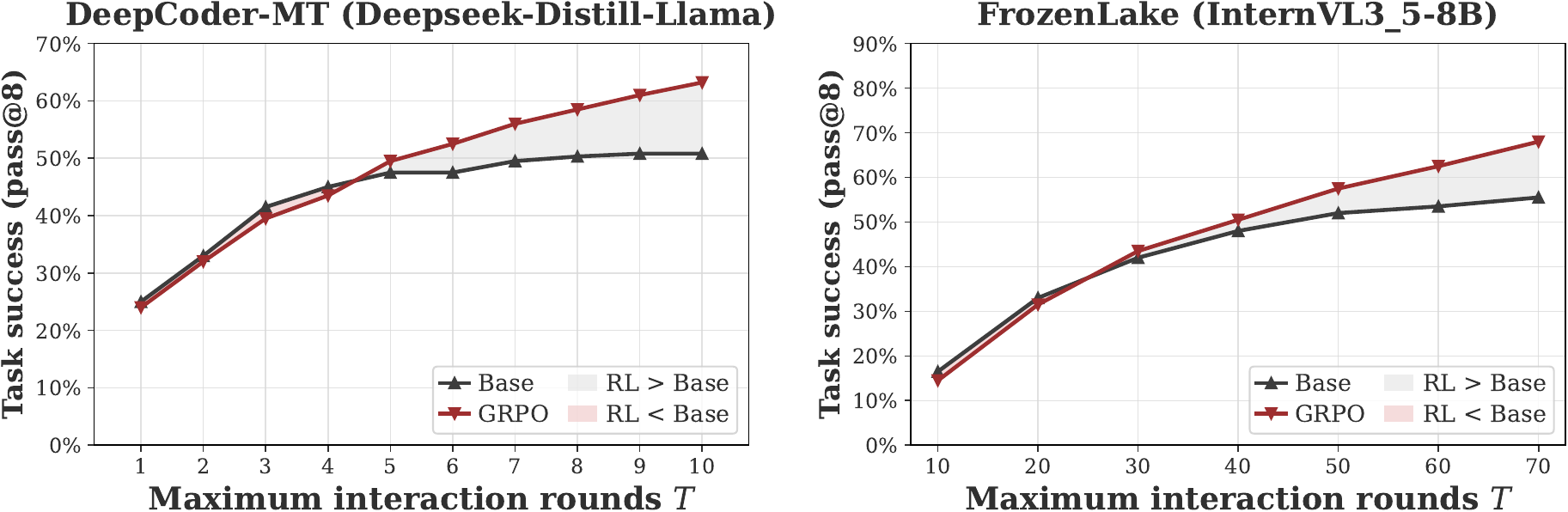}
    \caption{
        $\mathrm{Pass}@T$ results for InternVL3.5-8B on FrozenLake
        and DeepSeek-R1-Distill-Llama-8B on DeepCoder-MT.
    }
    \label{fig:pass_t_architectures}
\end{figure}

We again evaluate InternVL3.5-8B on FrozenLake and DeepSeek-R1-Distill-Llama-8B on DeepCoder-MT. As shown in Figure~\ref{fig:pass_t_architectures}, the $\mathrm{Pass}@T$ limitation persists across these architectures and is not specific to the Qwen3 family. As the interaction budget tightens, the advantage of RL-trained models diminishes and may disappear or reverse.

\subsection{Additional Tasks}
\label{app:pass_t_tasks}

\begin{figure}[t]
    \centering
    \includegraphics[width=0.9\textwidth]{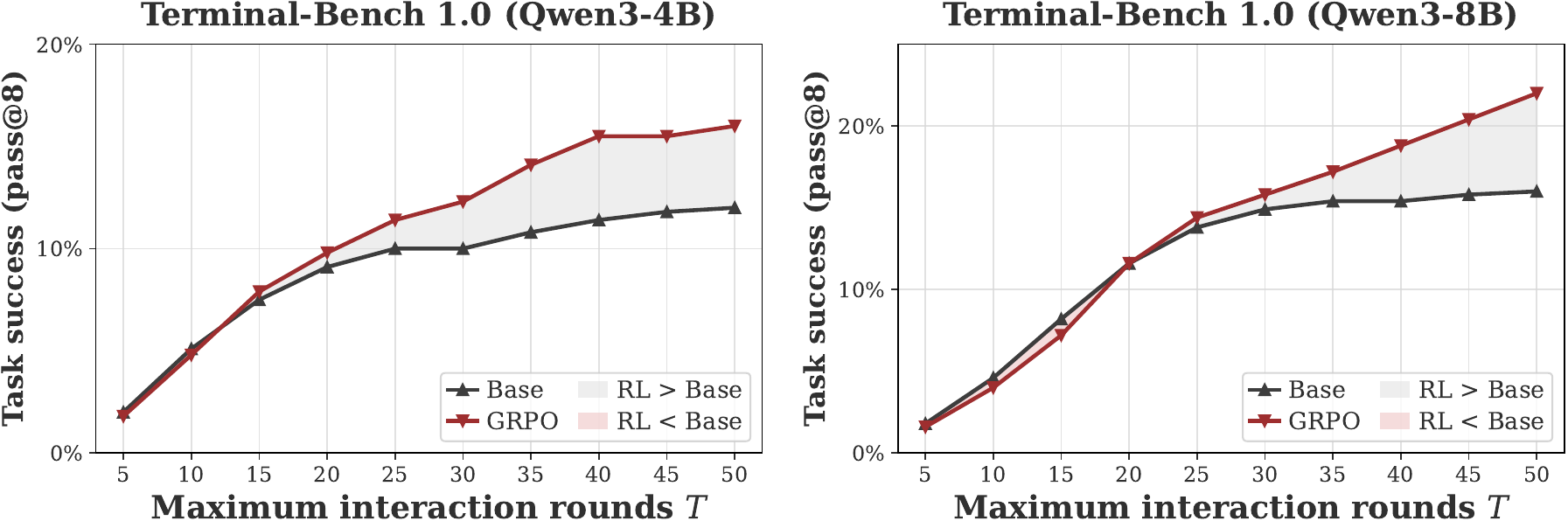}
    \caption{
        $\mathrm{Pass}@T$ results for the Qwen3 models on Terminal-Bench.
    }
    \label{fig:pass_t_terminal}
\end{figure}

As in Appendix~\ref{app:additional_pass_k}, we also evaluate the Qwen3 models on Terminal-Bench. Figure~\ref{fig:pass_t_terminal} shows that this limitation also arises in more realistic terminal tasks.
As turn budgets tighten, the gains of RL-trained models become inconsistent.
This further highlights the need for a comprehensive understanding of agentic RL to ensure stable and reliable real-world deployment.

Overall, the limitations and risks we identify recur across model architectures and tasks:
the advantage of agentic RL depends on the interaction budget and is not consistently maintained across budgets.
High-quality process rewards can mitigate this limitation, but such ideal supervision is not always available in real-world tasks.
This motivates our preliminary exploration of BudgetMix RL.

\clearpage

\section{Additional Results for Analysis Experiments}
\label{app:additional_analyses}

The preceding sections of the appendix focus on $\mathrm{Pass}@(K,T)$ results.
Here, we present additional results of analysis experiment to complement \S\ref{sec:step} and ~\ref{sec:mechanism} of the main text.

\begin{figure}[t]
\centering
\includegraphics[width=0.9\textwidth]{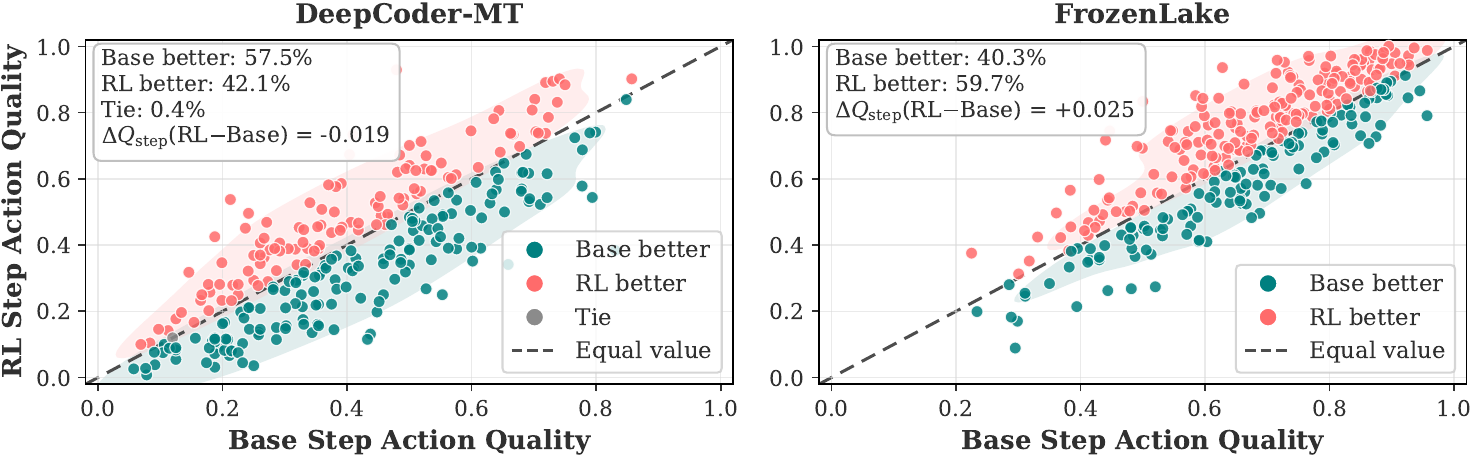}
\caption{Step-level action quality. Each point is one matched state; its coordinates are action quality of the base and RL model. Teal points favor the base model, red points favor the RL model.}
\label{fig:gigpo_matched}
\vspace{-5pt}
\end{figure}

\begin{figure}[t]
\centering
\includegraphics[width=0.9\textwidth]{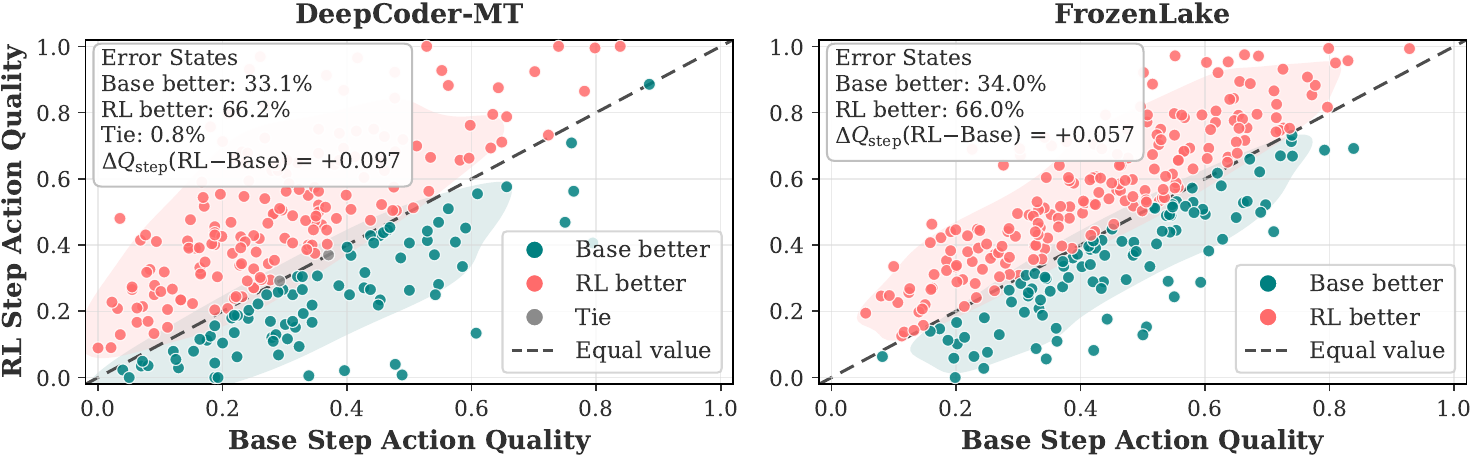}
\caption{Step-level action quality on error states. Each point is one error state.}
\label{fig:gigpo_error}
\vspace{-10pt}
\end{figure}

\begin{figure}[t]
\centering
\includegraphics[width=0.9\textwidth]{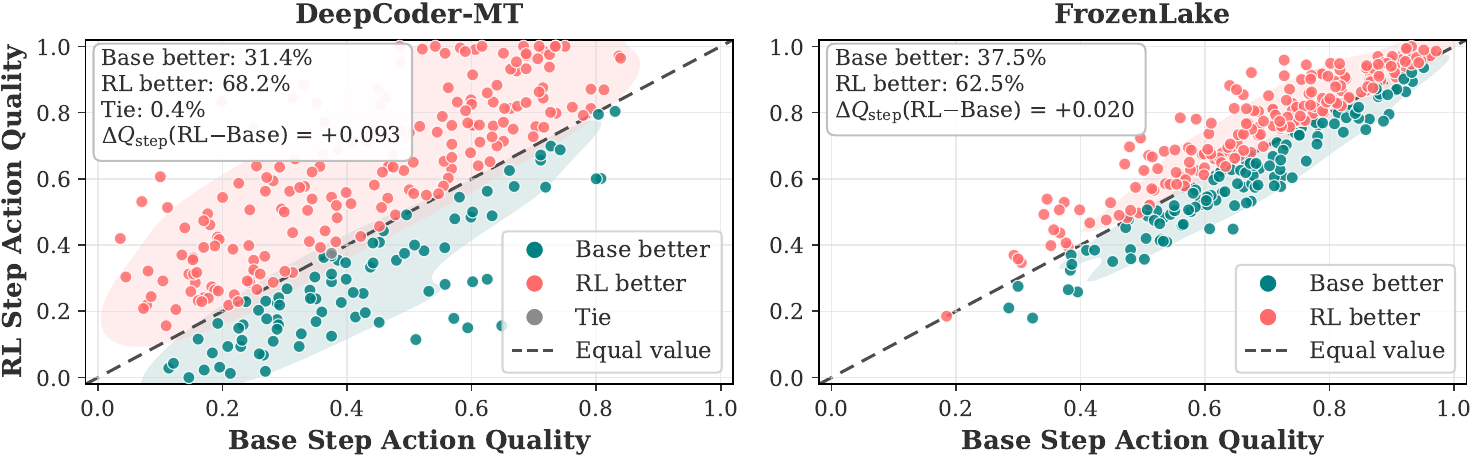}
\caption{Step-level action quality of \textbf{BudgetMix RL}.}
\label{fig:budgetmix_matched}
\vspace{-5pt}
\end{figure}

\begin{figure}[t]
\centering
\includegraphics[width=0.9\textwidth]{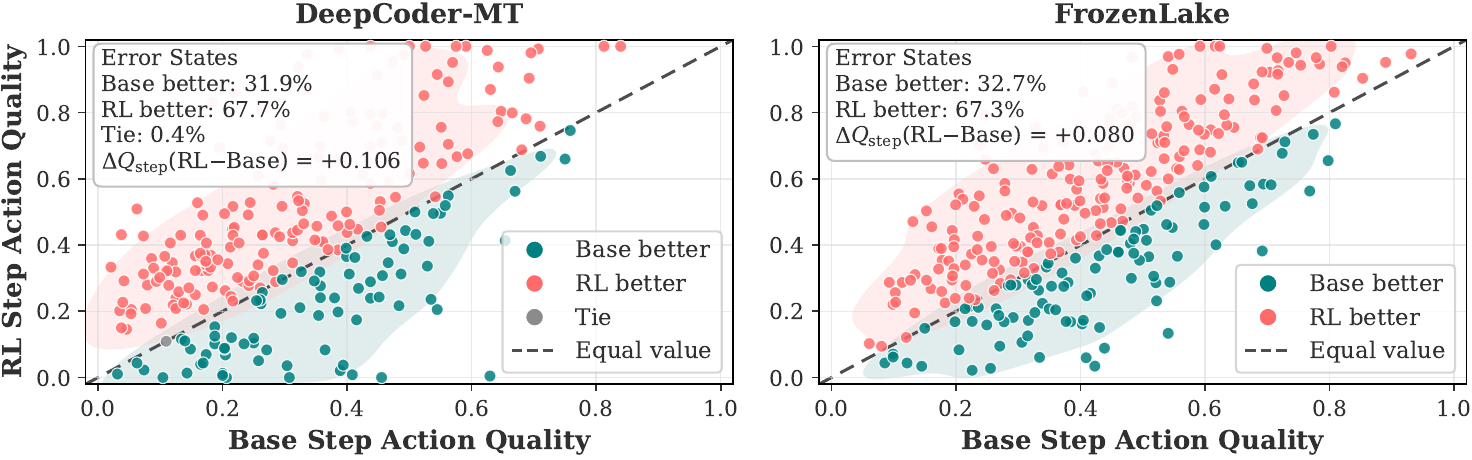}
\caption{Step-level action quality on error states of \textbf{BudgetMix RL}.}
\label{fig:budgetmix_error}
\vspace{-10pt}
\end{figure}

\subsection{Additional Results of Step-Level Action Quality}
\label{app:step_action_quality}

In \S\ref{sec:step} of the main text, we analyze how agentic RL affects single-step action quality. While the main experiments use GRPO, here we provide additional results with GiGPO. As shown in Figures~\ref{fig:gigpo_matched} and~\ref{fig:gigpo_error}, on FrozenLake, where GiGPO can obtain informative step-level credit signals, the RL model achieves better single-step action quality than the base model at both all matched states and error states. In contrast, on DeepCoder-MT, recurring matched states are difficult to obtain, making reliable step-level credit assignment substantially harder. Consequently, the RL model again does not consistently outperform the base model in single-step action quality. These results suggest that high-quality process-level supervision can improve single-step action quality. However, such supervision may be difficult to obtain in more realistic agentic tasks.

Similarly, we evaluate the single-step action quality of models trained with BudgetMix GRPO. As shown in Figure~\ref{fig:budgetmix_matched}, on randomly sampled matched states, BudgetMix RL yields a clear improvement in single-step decision quality.
Unlike standard GRPO, which does not consistently improve action quality and can even degrade it, as shown in Figure~\ref{fig:matched} of \S\ref{sec:step}, BudgetMix does not exhibit such degradation and instead produces consistently higher-quality local decisions.

\subsection{Additional Log-Likelihood Analysis}
\label{app:additional_likelihood}

\begin{figure}[t]
  \centering
  \includegraphics[width=0.9\textwidth]{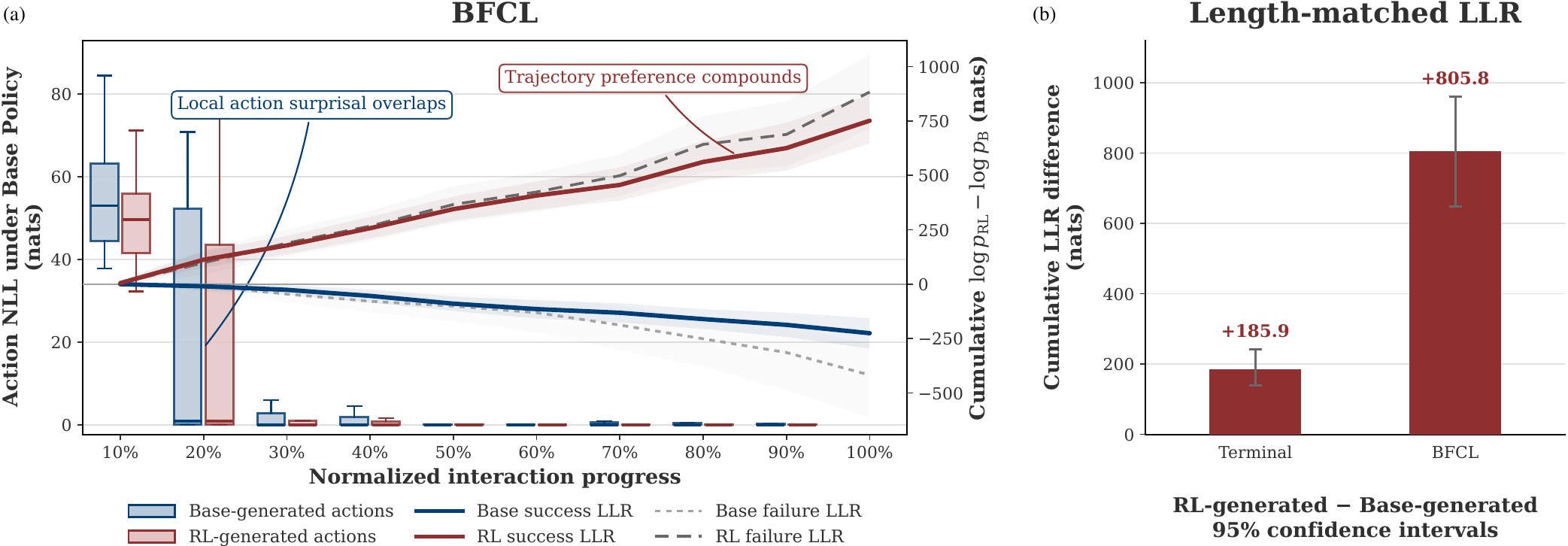}
    \caption{
        More log-likehood analysis of base and RL models. (a) Additional log-likelihood analysis on BFCL. (b) Cumulative LLR after length matching. Error bars indicate 95\% CI.
    }
    \label{fig:app_bfcl_likelihood}
\end{figure}

In \S\ref{sec:likelihood}, we present a mechanistic analysis based on log-likelihoods, primarily on Terminal-Bench. Here, we extend this analysis to BFCL. As shown in Figure~\ref{fig:app_bfcl_likelihood} (a), we observe a consistent pattern: base model assigns similar likelihoods to individual actions generated by base and RL models, yet at the trajectory level, base systematically assigns lower likelihoods to RL trajectories than to its own.  These results show a substantial shift between the RL and base policies over complete long-horizon trajectories, despite similar per-token NLLs under the base policy for RL and base actions.

\textbf{Controlling for output length.} We match base and RL trajectories within each task, requiring scored output token counts to differ by at most 10\% of the longer sequence. This yields 193 pairs across 89 tasks on Terminal and 283 pairs across 146 tasks on BFCL. Using the cumulative LLR from \S\ref{sec:likelihood} without token normalization, the matched difference between base and RL models is 185.9 nats on Terminal and 805.8 on BFCL, with both 95\% confidence intervals above zero (Figure~\ref{fig:app_bfcl_likelihood} (b)). The separation of cumulative LLRs thus helps us to partly rule out the influence of trajectory length.


\subsection{Additional Training Dynamics}
\label{app:additional_training_dynamics}

\begin{figure}[t]
    \centering
  \centering
  \includegraphics[width=0.8\textwidth]{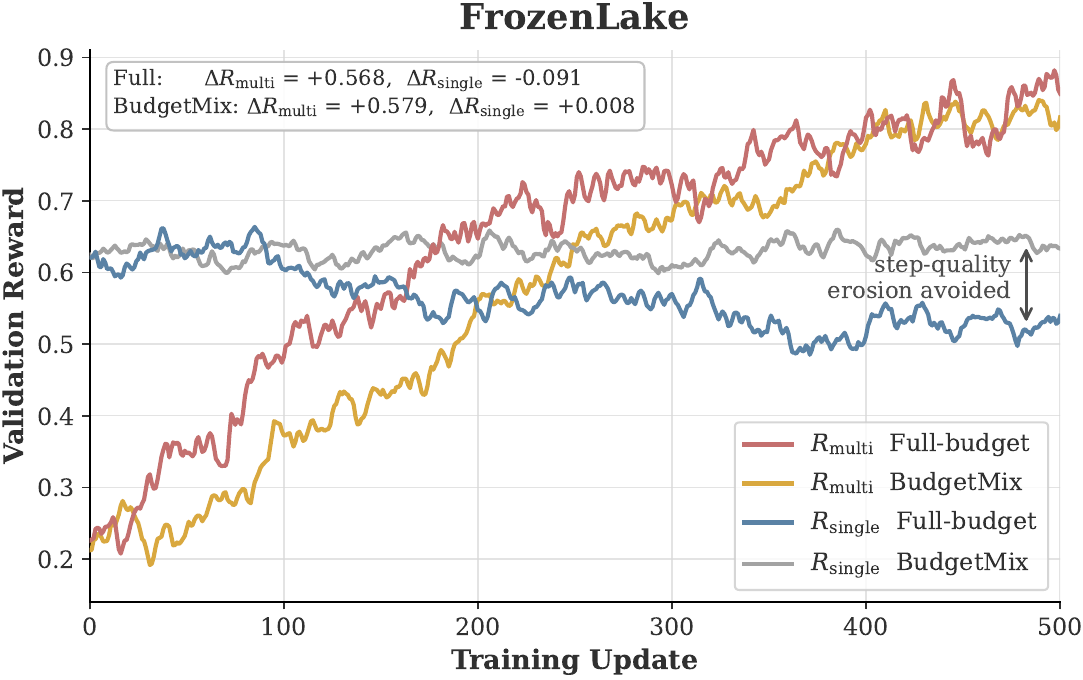}
  \caption{Training dynamics of the trajectory-level reward $R_{\mathrm{multi}}$ and step reward $R_{\mathrm{single}}$. }
  \label{fig:app_frozenlake_training_dynamics}
\end{figure}

In \S\ref{sec:dynamics}, we track how agentic RL affects trajectory success and single-step action quality as training progresses. Here, we provide additional training dynamics on FrozenLake. As shown in Figure~\ref{fig:app_frozenlake_training_dynamics}, under standard RL, trajectory-level reward $R_{\mathrm{multi}}$ (red curve) increases, indicating agent RL can improve trajectory success. However, single-step action quality reward $R_{\mathrm{single}}$ (blue curve) does not improve and instead tends to decline. Thus, trajectory-level gains during training do not necessarily translate into better local actions. However, with BudgetMix RL, single-step decision quality $R_{\mathrm{single}}$ (gray curve) remains stable or improves slightly, ending approximately $0.1$ above the standard RL curve. These results suggest that BudgetMix RL can help mitigate the decline in single-step decision quality. For each of FrozenLake and DeepCoder-MT, we construct a fixed pool of 512 states from base model trajectories on held-out tasks, using multiple rollout seeds and sampling across tasks and trajectory depths to encourage diversity. Each state includes its history and remaining budget. At each validation checkpoint, we sample 8 next-step actions per state and average their quality scores over actions and states to obtain $R_{\mathrm{single}}$. All checkpoints share the same state pool and scoring protocol.

\subsection{Additional Case Study Results}
\label{app:additional_case_study}

\begin{figure}[t]
    \centering
    \includegraphics[width=0.9\textwidth]{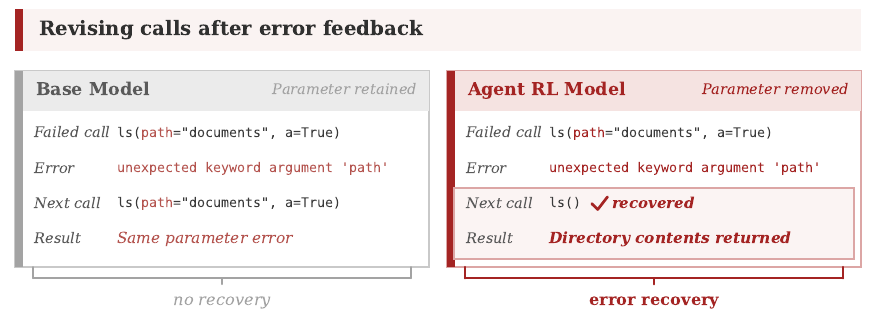}
    \caption{
        A BFCL example in which the RL model corrects an erroneous tool call.
    }
    \label{fig:bfcl_error_recovery_case}
\end{figure}

In \S\ref{sec:step}, we find that RL models produce higher-quality actions in error states, suggesting agent RL may improve error recovery.
Here, we provide a more direct case study on BFCL, examining whether models correctly revise a failed tool call in their next action. We collect trajectory prefixes containing tool-call errors (191), drawing half from RL rollouts (91) and half from base (100). Both models then take one action from each identical prefix. The RL model achieves an error correction rate of $41.4\%$, substantially exceeding the $27.2\%$ of base model. Figure~\ref{fig:bfcl_error_recovery_case} illustrates a case in which the RL model removes an unsupported argument rather than repeating the same error. Together, these results provide further support for our conclusion that agentic RL strengthens error recovery and resilience.

\end{document}